\pdfoutput=1

\documentclass[11pt]{article}

\usepackage[]{acl}

\usepackage{times}
\usepackage{latexsym}

\usepackage[normalem]{ulem}
\useunder{\uline}{\ul}{}
\usepackage{lscape}
\usepackage{longtable}

\usepackage[T1]{fontenc}

\usepackage[utf8]{inputenc}

\usepackage{microtype}

\usepackage{inconsolata}

\usepackage{multirow}
\usepackage{amsmath}
\usepackage{amsfonts}
\usepackage{adjustbox}
\usepackage{fancyvrb}
\usepackage{cprotect}
\usepackage[shortlabels]{enumitem}
\usepackage{graphicx}
\usepackage{float}
\usepackage{pdflscape}
\usepackage{longtable}
\usepackage{geometry}

\newcommand{\roberta}{\textsc{r}o\textsc{bert}a}
\newcommand{\robertabase}{\textsc{r}o\textsc{bert}a$_\textsc{base}$}
\newcommand{\scibert}{\textsc{scibert}}
\newcommand{\IndusBPE}{\textsc{IndusBPE}}

\newcommand{\llms}{\textsc{llm}s}
\newcommand{\nlp}{\textsc{nlp}}
\newcommand{\nasa}{\textsc{nasa}}
\newcommand{\cassiopea}{\textsc{indus}}
\newcommand{\cassbase}{\textsc{indus}$_\textsc{base}$}
\newcommand{\casssmall}{\textsc{indus}$_\textsc{small}$}
\newcommand{\cassretbase}{\textsc{indus}-\textsc{retriever}$_\textsc{base}$}
\newcommand{\cassretsmall}{\textsc{indus}-\textsc{retriever}$_\textsc{small}$}
\newcommand{\cassrank}{\textsc{indus}$_\textsc{ranker}$}
\newcommand{\climate}{\textsc{climate-change ner}}
\newcommand{\nasaqa}{\textsc{nasa-qa}}
\newcommand{\nasair}{\textsc{nasa-ir}}
\newcommand{\squad}{\textsc{sq}u\textsc{ad}}
\newcommand{\sde}{\textsc{sde}}
\newcommand{\ejclass}{$\textsc{ej}_\textsc{classifier}$}

\title{\cassiopea{}: Effective and Efficient Language Models for Scientific Applications}

\author{Bishwaranjan Bhattacharjee$^1$\thanks{Correspondence: \small  bhatta@ibm.com, mr0051@uah.edu, aashka.trivedi@ibm.com, rahul.ramachandran@nasa.gov}, Aashka Trivedi$^1$, Masayasu Muraoka$^1$, \\
\bf Muthukumaran Ramasubramanian$^3$, Takuma Udagawa$^1$, Iksha Gurung$^3$, \\
\bf Nishan Pantha$^3$, Rong Zhang$^1$, Bharath Dandala$^1$, Rahul Ramachandran$^2$,   \\
\bf  Manil Maskey$^2$,  Kaylin Bugbee$^2$, Mike Little$^4$, Elizabeth Fancher$^2$, Irina Gerasimov$^5$, \\
\bf Armin Mehrabian$^5$, Lauren Sanders$^6$, Sylvain Costes$^6$, Sergi Blanco-Cuaresma$^7$, \\
\bf Kelly Lockhart$^7$, Thomas Allen$^7$, Felix Grezes$^7$, Megan Ansdell$^8$,  Alberto Accomazzi$^7$,\\
\bf Yousef El-Kurdi$^1$,   Davis Wertheimer$^1$,  Birgit Pfitzmann$^{10}$\thanks{Work done while at IBM Research AI}, Cesar Berrospi Ramis$^1$, \\
\bf Michele Dolfi$^1$, Rafael Teixeira de Lima$^1$,    Panagiotis Vagenas$^1$,  S. Karthik Mukkavilli$^1$,\\
\bf  Peter Staar$^1$,  Sanaz Vahidinia$^8$, Ryan McGranaghan$^9$, Tsendgar Lee$^8$ \\ 
$^1$IBM Research AI,  $^2$NASA MFSC, $^3$UAH, $^4$ Navteca, $^5$ NASA GSFC, $^6$NASA Ames, \\ 
 $^7$Harvard-Smithsonian CfA,  $^8$NASA HQ, $^9$ JPL,  $^{10}$Smart City \& ERZ Zurich\\
}

\begin{document}
\maketitle
\begin{abstract}


Large language models (\llms{}) trained on general domain corpora showed remarkable results on natural language processing (\nlp{}) tasks. However, previous research demonstrated \llms{} trained using domain-focused corpora perform better on specialized tasks. Inspired by this insight, we developed \cassiopea{}, a comprehensive suite of \llms{} tailored for the closely-related domains of Earth science, biology, physics, heliophysics, planetary sciences and astrophysics, and trained using curated scientific corpora drawn from diverse data sources. The suite of models include: (1) an encoder model trained using domain-specific vocabulary and corpora to address \nlp{} tasks, (2) a contrastive-learning based text embedding model trained using a diverse set of datasets to address information retrieval tasks and (3) smaller versions of these models created using knowledge distillation for applications which have latency or resource constraints. We also created three new scientific benchmark datasets, \climate{} (entity-recognition), \nasaqa{} (extractive QA) and \nasair{} (IR) to accelerate research in these multi-disciplinary fields. We show that our models outperform both general-purpose (\roberta{}) and domain-specific (\scibert) encoders on these new tasks as well as existing tasks in the domains of interest. Furthermore, we demonstrate the use of these models in two industrial settings- as a retrieval model for large-scale vector search applications and in automatic content tagging systems.

\end{abstract}

\section{Introduction}

\begin{figure*}[t!]
\centering
\includegraphics[width=0.9\textwidth, height=6.5cm]{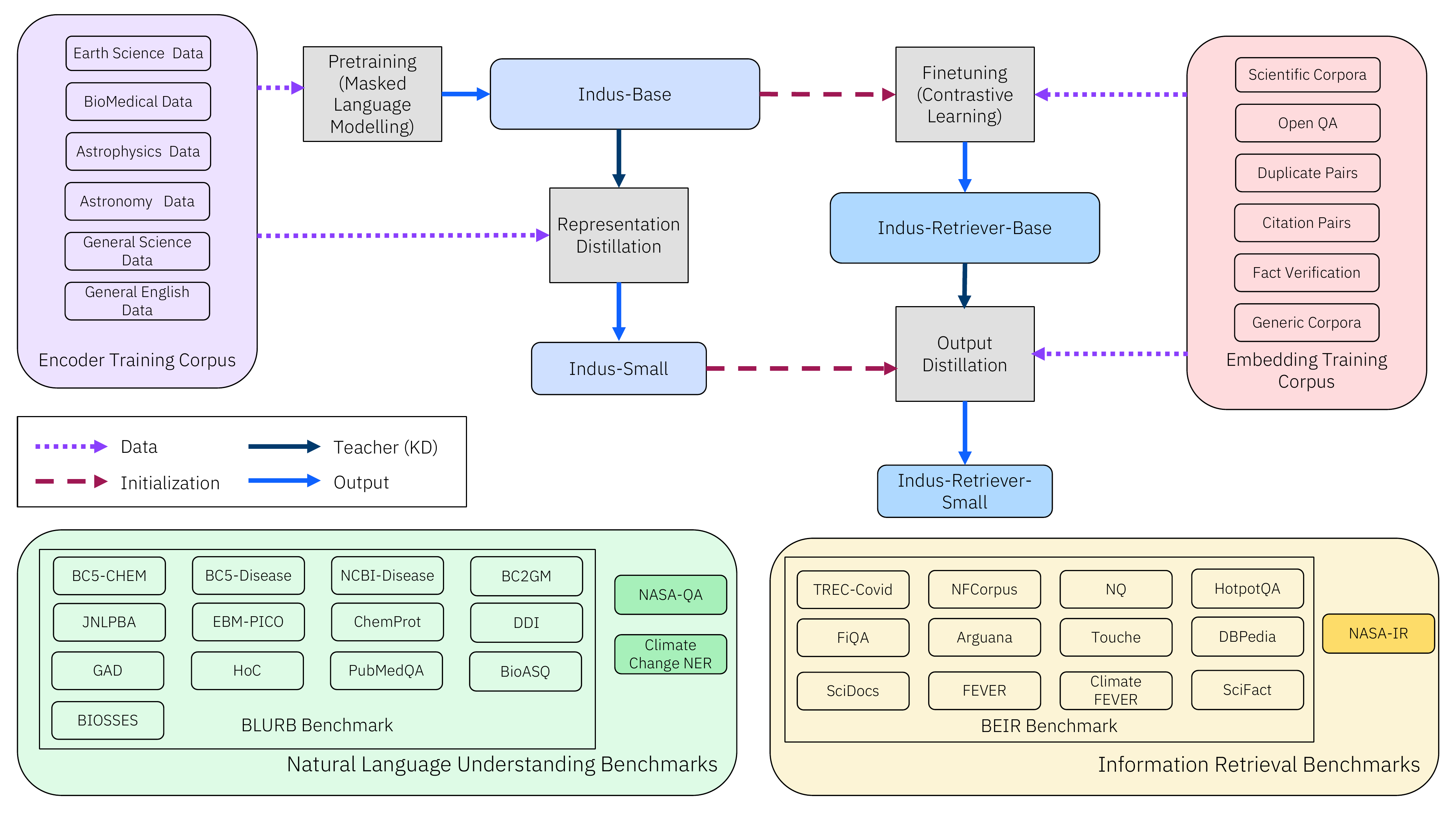}
\caption{Overview of \cassiopea{} models: the general-purpose encoder model and the retriever built from it, and their distilled counterparts. Also shown are the benchmarks used for evaluation, highlighting our new benchmarks, \nasaqa{}, \climate{} and \nasair{}.}
\label{fig-intro}
\end{figure*}  

Large language models (\llms{}) trained on huge amounts of data have demonstrated impressive capabilities on natural language understanding and generation tasks. Most popular \llms{} rely on the transformer architecture \cite{vaswani_attention} and are trained using general-purpose corpora like Wikipedia or CommonCrawl \cite{devlin-etal-2019-bert, liu2019roberta, lewis-etal-2020-bart, raffel-2020-t5, brown-2020-gpt, touvron2023llama}. Although these general-purpose models exhibited strong performance, the distributional shift of vocabulary led to sub-optimal performance on domain-specific natural language understanding and generation tasks \cite{beltagy-etal-2019-scibert}. Following this observation, several domain-specific \llms{} like \textsc{scibert} \cite{beltagy-etal-2019-scibert}, \textsc{biobert} \cite{lee-2019-biobert}, \textsc{matbert} \cite{walker2021impact}, \textsc{batterybert} \cite{huang2022batterybert} and  \textsc{scholarbert} \cite{hong2023diminishing} were developed to improve accuracy on in-domain \textsc{nlp} tasks.

In this research, we specifically focused on inter-disciplinary scientific topics related to astrophysics, physics, Earth science, heliophysics, planetary sciences and biology. While the training corpora of existing domain-specific models such as \textsc{scibert}, \textsc{biobert} and \textsc{scholarbert} partially cover some of these fields, there is no model available that encompasses all of the fields of interest collectively. 

Thus, we developed \cassiopea{}, a collection of encoder-based \llms{} focused on these domains of interest (Figure \ref{fig-intro}) trained using curated corpora from diverse sources. Specifically, we make the following contributions:

\begin{enumerate}[topsep=0pt, itemsep=0pt, leftmargin=.2in, parsep=0pt]
    \item Utilizing the byte-pair encoding (BPE) algorithm, we constructed \IndusBPE{}, a customized tokenizer from the curated scientific corpus.
    \item We pretrained \textbf{encoder-only \llms{}} using curated scientific corpora and the \IndusBPE{} tokenizer (\S\ref{sec:data}, \S\ref{sec:encoders}). We further created \textbf{sentence-embedding models} by fine-tuning the encoder-only models with a contrastive learning objective (\S\ref{sec:embeddings}).
    We also trained \textbf{smaller, efficient versions} of these models using distillation.
    \item We created \textbf{three new scientific benchmark datasets}, \climate{} (an entity recognition task), \nasaqa{} (an extractive question answering task) and \nasair{} (a retrieval task) (\S\ref{sec:benchmarks}) to further accelerate research in this multi-disciplinary field.
    \item We demonstrate strong performance by our models on these benchmark tasks as well as on existing domain-specific benchmarks, outperforming general-purpose models like \roberta{} \cite{liu2019roberta} as well as scientific-domain encoders like \scibert{} \cite{beltagy-etal-2019-scibert}. We also show that the knowledge-distilled models achieved a significant reduction in latency while maintaining strong performance compared to the original models on most of these tasks.
    \item We describe two industrial application areas of \cassiopea{} models in the scientific domain, where they outperform existing general-purpose models.
  
\end{enumerate}

\section{Data}
\label{sec:data}

Sufficient high-quality in-domain corpora is essential to develop models that perform better than their counterparts trained on open-domain corpora. We meticulously identified corpora for each of the aforementioned domains, and created English-only models for containment. 
Specifically, for each domain, we used open-source data which has a permissive license, and further augmented them with full text papers and material contributed by providers mentioned below.
We now briefly describe each data source, and present statistics of the data in Table \ref{tab:pretrain_ds}.


\begin{table}[t!]
\centering
\tabcolsep 2pt
\scalebox{0.8}{
\begin{adjustbox}{max width=0.47\textwidth}
\begin{tabular}{lcrr}
\hline
\multicolumn{1}{c}{Dataset} & \multicolumn{1}{c}{Domain} & \multicolumn{1}{c}{\#Tokens} & \multicolumn{1}{c}{Ratio} \\
\hline
\nasa{} \textsc{cmr} & Earth Science & 0.3B & 1\% \\
\textsc{ams} and \textsc{agu} papers & Earth Science & 2.8B & 4\% \\
English Wikipedia & General & 5.0B & 8\% \\
PubMed Abstracts & Biomedical & 6.9B & 10\% \\
\textsc{pmc} & Biomedical & 18.5B & 28\% \\
\textsc{sao}/\nasa{} \textsc{ads} & Astronomy,  & 32.7B & 49\% \\
& Astrophysics, & &\\
& Physics, & &\\
& General Science & &\\
\textbf{Total} & & 66.2B & 100\% \\
\hline
\end{tabular}
\end{adjustbox}}
\caption{Basic statistics of our pretraining dataset.}
\label{tab:pretrain_ds}
\end{table}

\begin{itemize}[topsep=0pt, itemsep=0pt, leftmargin=.2in, parsep=0pt]
    \item \textbf{\textsc{sao}/\nasa{} Astrophysics Data System (\textsc{ads})\footnote{https://ui.adsabs.harvard.edu}}:  The biggest source of data, covering publications in astronomy and astrophysics, physics and general science including all arXiv e-prints.
    \item \textbf{PubMed Central (\textsc{pmc})\footnote{https://www.ncbi.nlm.nih.gov/pmc}} : An archive of biomedical and life science journal literature maintained by National Library of Medicine and National Institutes of Health. We used the portion of \textsc{pmc} that has a commercial-friendly license, along with the PubMed abstracts of all the articles in \textsc{pmc}.
    \item \textbf{American Meteorological Society (\textsc{ams})}\footnote{https://www.ametsoc.org/index.cfm/ams/publications/}:  We used full-text journal documents spanning topics in Earth systems, Earth interactions, applied meteorology, climatology, physical oceanography, atmospheric sciences, climate, hydrometeorology, weather, forecasting, and societal impacts.
    \item \textbf{American Geophysical Union (\textsc{agu})\footnote{https://agupubs.onlinelibrary.wiley.com/}}:  Journal documents on the topics of atmospheres, biogeosciences, Earth surface, machine learning and computation, oceans, planets, solid Earth, and space physics. 
    \item \textbf{\nasa{} Common Metadata Repository (\textsc{cmr})\footnote{https://www.earthdata.nasa.gov/eosdis/science-system-description/eosdis-components/cmr}}: A high-quality, continuously evolving metadata system that catalogs all data and service metadata records for \nasa{}'s Earth Science Data and Information System. 
\end{itemize}

\section{Methodology: Encoder Models}
\label{sec:encoders}

\paragraph{\IndusBPE{} Tokenizer}
\label{sec:tokenizer}
We trained an uncased BPE tokenizer \cite{radford2019gpt2}, \IndusBPE{}, using a subset of our training dataset (\S\ref{sec:data}). We set the vocabulary size to $50265$ (equal to that of the \roberta{} tokenizer \cite{liu2019roberta}).

We performed a brief analysis of the differences between the vocabularies of \IndusBPE{} and the \roberta{} tokenizer. Out of $50265$ tokens, $44.5\%$ tokens are common in both tokenizers while the remaining $55.5\%$ tokens are included only in either tokenizer, indicating a significant distributional shift in domain. To further understand this effect,  we applied both tokenizers on $1000$ randomly sampled text fragments from our datasets. As shown in Table \ref{tab:token_count}, \IndusBPE{} tokenizer produced fewer tokens than the \roberta{} tokenizer, leading to an 8\% drop in computation cost during training. 



\begin{table}[t!]
\centering
\scalebox{0.9}{
\begin{adjustbox}{width=0.45\textwidth}
\begin{tabular}{lrrr}
\hline
Tokenizer & \multicolumn{1}{c}{ADS} & \multicolumn{1}{c}{PMC} & \multicolumn{1}{c}{Wikipedia} \\
\hline
\roberta{} & 12,867,439 & 7,549,075 & 15,859 \\
\ +lower\_cased & 12,862,227 & 7,557,868 & 16,901 \\
\IndusBPE{} & 12,309,023 & 6,920,659 & 16,056 \\
\hline
\end{tabular}
\end{adjustbox}}
\caption{Number of tokens produced by \roberta{} and \IndusBPE{} tokenizers on 1k samples from each dataset. Fewer tokens lead to a smaller computation cost.}
\label{tab:token_count}
\end{table}

\paragraph{Encoder Model}
\label{sec:encoder}
We trained \cassbase{}\footnote{https://huggingface.co/nasa-impact/nasa-smd-ibm-v0.1} using a masked language modeling objective. The model architecture follows \robertabase{} \cite{liu2019roberta}, with 12 layers and 125M parameters. 

\paragraph{Knowledge Distillation for Efficient Encoder Model}
\label{sec:kd-encoder}
We also trained a smaller model, \casssmall{}\footnote{https://huggingface.co/nasa-impact/nasa-smd-ibm-distil-v0.1}, with 38M parameters through knowledge distillation using \cassbase{} as the teacher. \casssmall{} follows a 4-layer architecture recommended by the Neural Architecture Search engine \cite{trivedi2023neural} with an optimal trade-off between performance and latency. We adopted the distillation objective proposed in MiniLMv2 \cite{wang-etal-2021-minilmv2} to transfer fine-grained self-attention relations, which has been shown to be the current state-of-the-art \cite{udagawa-etal-2023-comparative}.

\section{Methodology: Sentence Embedding Models}
\label{sec:embeddings}

Sentence embedding models represent text as low-dimensional vectors for efficient use in dense retrieval systems, such as Retrieval Augmented Generation, where relevant passages for a query are identified by the similarity between their embeddings \cite{karpukhin-etal-2020-dense}. Embedding models are trained using a contrastive learning objective \cite{khosla_contrastive_2020, gao-etal-2021-simcse}, which pushes the embeddings of a query closer to those of relevant passages and further away from those of non-relevant ones. We use the improved contrastive loss proposed in \citet{li2023general} which introduces an additional bidirectional signal to expand negatives.

\paragraph{Base Embedding Model}  We created our sentence embedding model, \cassretbase{}\footnote{https://huggingface.co/nasa-impact/nasa-smd-ibm-st-v2}, by fine-tuning \cassbase{}, following a bi-encoder framework \cite{reimers-gurevych-2019-sentence}. Similar to prior work \cite{wang2022text, li2023general, xiao2023cpack}, we employed a stage-wise training approach. We first train on a large corpus of naturally occurring pairs collected from internet sources, and specifically include data from the science domain. Furthermore, we created a domain-specific dataset from the \textsc{ads} data (\S\ref{sec:data}) by including title-abstract pairs. Then, we finetune on high quality annotated datasets (e.g., question-answer pairs). Appendix \ref{app:embedding-data} contains comprehensive details about the datasets used in training. For both stages, we used large batch sizes and in-batch negatives to better approximate the contrastive objective.

\paragraph{Knowledge Distillation for Embedding Model} To optimize the latency for retrieval applications, we also created a small retriever model, \cassretsmall{}\footnote{https://huggingface.co/nasa-impact/nasa-ibm-st.38m}, with the aim to transfer the capability of the large teacher model (\cassretbase{}, with 12 layers and an embedding dimension of 768) to smaller student model (\casssmall{}, with 4 layers and an embedding dimension of 576), by distilling the teacher's distribution of similarity scores. Specifically, we use the distillation loss described in \citet{xu-etal-2023-distillcse}

Here, we find it beneficial to first conduct an embedding-oriented pretraining step, as presented in Retro-MAE \cite{xiao-etal-2022-retromae}, on about 56M sentences from English Wikipedia, BooksCorpus, and StackExchange data. We observed that this step is not necessary in the larger model, but provides significant improvement in the smaller one. For distillation, we found that a stage-wise training approach does not benefit performance (ablation presented in Appendix \ref{app:embedding-stagewise-ablation}). We thus distilled in a single step with all the data described in Appendix \ref{app:embedding-data}, also adding labelled pairs from \textsc{fever} \cite{thorne-etal-2018-fever} and \textsc{hotpotqa} \cite{yang-etal-2018-hotpotqa}.

\section{Creating Benchmarks}
\label{sec:benchmarks}
Benchmark datasets play a crucial role in assessing the language understanding capabilities of models. However, there is an absence of datasets tailored for the diverse and multidisciplinary fields under study. Thus, to effectively benchmark the proposed \nlp{} models, 
we introduced three new datasets for NER, QA and IR. Appendix \ref{app:benchmark-details} compares the sizes of these datasets to popularly used benchmarks.

\subsection{\climate{}}
\label{sec:climate_change_ner}

\begin{table}[t!]
\centering
\small
\begin{adjustbox}{width=0.45\textwidth}
\begin{tabular}{lccc}
\hline
& Train & Validation & Test \\
\hline
Num. Abstracts & 382 & 77 & 75 \\
Num. Tokens & 32,031 & 6,443 & 5,850 \\
\hline
\multicolumn{4}{l}{Entity Labels} \\
\multicolumn{4}{l}{\textit{climate-nature, climate-greenhouse-gases, climate-assets,}} \\
\multicolumn{4}{l}{\textit{climate-problem-origins, climate-mitigations, }} \\
\multicolumn{4}{l}{\textit{climate-properties, climate-impacts, climate-datasets, }} \\
\multicolumn{4}{l}{\textit{climate-organizations, climate-observations, }} \\
\multicolumn{4}{l}{\textit{climate-models, climate-hazards, climate-organisms}}\\
\hline
\end{tabular}
\end{adjustbox}
\caption{\climate{} statistics and entities.}
\label{tab:climate_change_stats}
\end{table}

\climate{}\footnote{https://huggingface.co/datasets/ibm/Climate-Change-NER} focuses on understanding and addressing climate-related topics across various domains. This 
comprises 534 abstracts sourced from Semantic Scholar Academic Graph \cite{kinney-etal-2023-s2odp}, collected using a seed set of climate-related keywords such as \textit{wildfire} or \textit{floods}.
The abstracts were annotated 
with entities of interest that originate from complex taxonomies used in climate-related literature as shown in Table \ref{tab:climate_change_stats}. 

\begin{table*}[h]
\centering
\scalebox{0.85}{
\begin{adjustbox}{width=1\textwidth}
\begin{tabular}{ccc|rrrrr|rrr}
\hline
\multicolumn{1}{c}{} & \multicolumn{1}{c}{} & \multicolumn{1}{c|}{} &
\multicolumn{5}{c|}{Base model (125M params.)} &
\multicolumn{3}{c}{Small model ($\sim$30M params.)} \\

\multicolumn{1}{c}{Task} & \multicolumn{1}{c}{Metric} & \multicolumn{1}{c|}{Dataset} &
\multicolumn{1}{c}{\roberta} &
\multicolumn{1}{c}{\textsc{Sci}\textsc{Bert}} &
\multicolumn{1}{c}{\textsc{PubMed}} &
\multicolumn{1}{c}{\textsc{BioLink}} &
\multicolumn{1}{c}{\cassbase} &
\multicolumn{1}{|c}{\textsc{TinyBERT}} &
\multicolumn{1}{c}{\textsc{MiniLM}} &
\multicolumn{1}{c}{\casssmall{}} \\

\hline

\multirow{5}{*}{NER} & \multirow{5}{*}{Entity F1} & BC5-chem &
90.3 (0.2) & 91.4 (0.2) & 93.2 (0.1) & \textbf{93.3} (0.2) & \textbf{93.3} (0.2) & 
84.6 (0.2) & 86.1 (0.3) & \textbf{90.7} (0.1) \\

& & BC5-disease &
81.5 (0.3) & 83.7 (0.3) & \textbf{85.4} (0.3) & 85.3 (0.3) & 85.2 (0.3) & 
74.0 (0.4) & 77.4 (0.3) & \textbf{81.3} (0.3) \\

& & NCBI-disease &
87.6 (0.6) & 87.6 (0.4) & 88.2 (0.6) & 88.2 (0.5) & \textbf{88.3} (0.4) & 
81.2 (0.4) & 83.1 (0.5) & \textbf{85.6} (0.6) \\

&  & BC2GM &
82.1 (0.3) & 82.3 (0.2) & 84.3 (0.3) & \textbf{84.7} (0.2) & 84.0 (0.3) & 
74.7 (0.4) & 77.1 (0.2) & \textbf{79.7} (0.3) \\

& & JNLPBA &
79.1 (0.2) & 78.2 (0.2) & 79.3 (0.2) & 78.9 (0.2) & \textbf{80.3} (0.2) & 
70.3 (0.2) & 73.4 (0.3) & \textbf{75.7} (0.2) \\

\hline

PICO & Macro F1 & EBM PICO &
72.3 (0.3) & 72.4 (0.3) & 72.9 (0.3) & \textbf{73.4} (0.2) & 73.1 (0.2) & 
67.4 (0.2) & 70.3 (0.1) & \textbf{73.1} (0.2) \\

\hline

\multirow{3}{*}{\shortstack[c]{Relation\\ Extraction}} & \multirow{3}{*}{Micro F1} & ChemProt &
50.4 (28.2) & 73.9 (0.7) & 77.2 (0.6) & \textbf{77.9} (0.4) & 76.9 (0.5) &
56.2 (3.2) & 55.9 (2.1) & \textbf{71.7} (0.9) \\

& & DDI &
78.6 (1.5) & 80.1 (1.0) & 80.6 (1.1) & 81.2 (0.6) & \textbf{81.7} (0.5) & 
39.3 (5.3) & 51.5 (2.9) & \textbf{69.0} (1.2) \\

& & GAD &
80.0 (1.1) & 81.6 (1.2) & \textbf{82.4} (1.2) & 82.1 (1.5) & 79.4 (5.6) & 
76.4 (1.3) & 77.3 (1.0) & \textbf{81.3} (0.7) \\

\hline

\shortstack[c]{Document\\ Classification} & Micro F1 & HoC &
82.2 (0.7) & 83.1 (0.6) & \textbf{84.5} (0.4) & 84.4 (0.5) & 83.7 (0.5) & 
41.6 (6.8) & 62.8 (4.7) & \textbf{80.2} (0.6) \\

\hline

\multirow{2}{*}{\shortstack[c]{Question\\ Answering}} & \multirow{2}{*}{Accuracy} & PubMedQA &
53.1 (3.3) & 54.3 (3.8) & 55.2 (5.5) & \textbf{59.1} (6.2) & 58.2 (6.7) & 
50.3 (1.4) & 51.6 (1.7) & \textbf{56.1} (1.4) \\

& & BioASQ &
69.1 (4.8) & 74.6 (4.5) & 84.3 (5.5) & \textbf{84.9} (10.5) & 69.6 (5.8) & 
74.3 (3.6) & 66.7 (2.3) & \textbf{75.4} (3.3) \\

\hline

\shortstack[c]{Sentence\\ Similarity} & Pearson & BIOSSES &
79.8 (6.3) & 86.3 (3.5) & \textbf{92.2} (1.1) & 91.1 (2.6) & 72.2 (9.5) & 
\textbf{88.2} (1.1) & 26.6 (8.7) & 70.4 (3.3) \\

\hline

\textbf{Micro Average} & - & - &
75.9 (3.7) & 79.2 (1.3) & 81.5 (1.3) & \textbf{81.9} (1.8) & 78.9 (2.4) & 
67.6 (1.9) & 66.1 (1.9) & \textbf{76.2} (1.0) \\

\textbf{\shortstack[c]{Macro Average}} & - & - &
74.9 (3.7) & 78.2 (1.6) & 80.9 (1.4) & \textbf{81.2} (3.9) & 76.4 (3.2) & 
65.6 (2.4) & 60.6 (3.0) & \textbf{74.3} (1.3) \\

\hline

\end{tabular}
\end{adjustbox}}
\caption{Evaluation on BLURB. Standard deviation across 10 random seeds in parenthesis. Macro avg. reported across datasets and micro avg. computed by averaging scores on each task then averaging across task averages.}

\label{tab:res_blurb_full1}
\end{table*}

\subsection{\nasaqa{}}

We created \nasaqa{}\footnote{https://huggingface.co/datasets/nasa-impact/nasa-smd-qa-benchmark}, an extractive QA benchmark dataset focused on the Earth science domain (\textsc{es}). Specifically, we sourced 39 paragraphs from \textsc{es} papers appearing in \textsc{agu} and \textsc{ams} journals (\S\ref{sec:data}), and subject matter experts formulated questions and annotated the spans of the paragraph that contain the answer. We used 29 paragraphs (145 questions) as the training set and remaining 10 paragraphs (50 questions) for evaluation. The training set was further augmented with paragraphs and QA pairs related to \textsc{es} (oxygen, amazon rain forest and geology) from the \squad{} dataset \cite{sq2}. This resulted in a training set comprising 686 paragraphs with 5,081 questions (2,817 answerable and 2,264 unanswerable).

\subsection{\nasair{}}
Finally, we constructed a domain-specific information retrieval benchmark dataset, \nasair{}\footnote{https://huggingface.co/datasets/nasa-impact/nasa-smd-IR-benchmark}, spanning almost 500 QA pairs related to the Earth science, planetary science, heliophysics, astrophysics and biological physical sciences domains. We sampled a set of 166 paragraphs from \textsc{agu, ams, ads, pmc} and PubMed (\S\ref{sec:data}) and manually annotated them with 3 questions that are answerable from each of these paragraphs, resulting in 498 questions (398 questions in the test set and 100 in the validation set- this test is designed to be evaluated in a zero shot fashion). We also sampled random abstracts from \textsc{ads} to enhance our corpus. Each question has only one relevant document, and we use the Recall@10 evaluation metric.

\section{Experimental Results}
\label{sec:exp_res}

\paragraph{Baselines} We compared \cassiopea{} models against open source models of similar sizes (all models obtained from HuggingFace):
\begin{itemize}[topsep=0pt, itemsep=0pt, leftmargin=.2in, parsep=0pt]
\item \cassbase{} was compared to \robertabase{}, \scibert{}, \textsc{PubMedBERT}, and \textsc{BioLinkBert}.

\item \casssmall{} was compared to \textsc{minilm} (6-layer) and \textsc{tinybert} (4-layer). 

\item \cassretbase{} was compared to \textsc{bge}$_\textsc{base}$\ and a \robertabase{} model finetuned with the same method presented in \S\ref{sec:embeddings}.

\item \cassretsmall{} was compared to \textsc{minilm-v2} and \textsc{bge}$_\textsc{small}$. 

\end{itemize}

\subsection{Natural Language Understanding Benchmarks}

\subsubsection{BLURB}
We evaluated our models on BLURB \cite{gu2022blurb}, a benchmark suite for natural language understanding and reasoning tasks in the biomedical domain. We followed the original work to compute the overall score (i.e., macro average).

Table~\ref{tab:res_blurb_full1} shows the evaluation results. Among base models, \cassbase{} significantly outperformed the general-purpose \roberta{} model while achieving competitive performance to the bio-domain-specific models, namely \textsc{scibert}, \textsc{PubMedBERT}, and \textsc{BioLinkBert}, in which the Macro Average of our model is still within two standard deviations ($76.4+3.2*2=82.8$), thus, the differences are not statistically significant. For smaller models, we noticed \casssmall{} outperformed the baselines, \textsc{tinybert} and \textsc{minilm}, by a large margin in most cases, showing significant difference from second best models in NER, PICO, relation extraction, and document classification tasks. This demonstrates the effectiveness of knowledge distillation from our domain-specific teacher model, \cassbase{}.

We noticed domain specific large baseline models tend to perform better than our model on paired input-text tasks, such as QA and semantic similarity tasks, although the results have relatively large standard deviations. We hypothesize that pre-training with paired texts in \textsc{bert}-style models (e.g., \textsc{scibert} and \textsc{PubMedBERT}) in contrast to the \roberta{}-style models (e.g., \roberta{} and \cassiopea{}) may be beneficial for such paired input-text tasks. This is consistent with the observations of \citet{tinn2023fine}\footnote{Specifically, as noted in their paper,``\textit{pretraining with single sequences leads to a substantial performance drop in the sentence similarity task. ... therefore pretraining with 2 text segments helps.}''}.

\subsubsection{\climate{}}

\begin{table}[t]
\centering
\scalebox{0.7}{\begin{tabular}{lcc}
\hline
Model & \climate{} & \nasaqa{} \\
& F1 (SD) & F1 (SD)\\
\hline
\roberta{} & 60.8 (0.8) & 66.8 (3.1) \\
\scibert{} & 61.8 (0.7)  & 63.5 (1.9)\\
\cassbase{} & \textbf{64.0} (1.0) & \textbf{68.2} (2.9) \\
\hline
\textsc{tinybert} & 34.3 (1.6) & 43.2 (2.3) \\
\textsc{minilm} & 44.7 (1.3) & \textbf{59.2} (3.9)\\
\casssmall{} & \textbf{54.8} (0.8) & 47.4 (1.8) \\
\hline
\end{tabular}}
\caption{\climate{} and \nasaqa{} benchmark results. Standard deviation  for \climate{} over 10 random seeds and \nasaqa{} over 3 random seeds in parenthesis.}
\label{tab:climate_change_esqa}
\end{table}

As shown in Table~\ref{tab:climate_change_esqa}, our models clearly outperformed the corresponding baseline models on the \climate{} task, suggesting the effectiveness of training on large domain-specific data.

\subsubsection{\nasaqa{}}

As mentioned in \S\ref{sec:benchmarks}, we augmented the training set with relevant \squad{} pairs for fine-tuning. All models are fine tuned for 15 epochs, and the results are shown in Table~\ref{tab:climate_change_esqa}. We observed that \cassbase{} outperformed all models of similar sizes, while \casssmall{} had relatively strong performance compared to its counterparts.

\begin{table}[t!]
\centering
\begin{adjustbox}{width=0.48\textwidth}
\begin{tabular}{l|c|c|c}
\hline
Model & \nasair{} $\uparrow$ & BEIR Avg. $\uparrow$  & Retrieval Time $\downarrow$ \\
& (Recall@10) & (NDCG@10) &  (s) \\
 \hline
\robertabase{} &  0.66 & 0.37 & 1.20 \\
\textsc{bge}$_\textsc{base}$ &  0.67 &  \textbf{0.52} &  1.18 \\
\cassretbase{} &  \textbf{0.71} & 0.41 & 1.19 \\
\hline
\textsc{minilm-v2} &   0.62 &  0.39 & 0.24 \\
\textsc{bge}$_\textsc{small}$ & 0.66&  \textbf{0.51} & 0.42 \\
\cassretsmall{} & \textbf{0.73} & 0.42 & 0.26 \\
\hline
\end{tabular}
\end{adjustbox}
\caption{Evaluation results on \nasair{} and BEIR, and average retrieval time per query on the NQ test set on an A100 GPU. Retrieval time includes time to encode the query \& corpus and time to retrieve relevant documents.
}
\label{tab:res_retrieval}
\end{table}
    
\subsection{Information Retrieval Benchmarks}

We evaluated our models on the \nasair{} dataset as well as BEIR Benchmark \cite{n2021beir}, which consists of 12 retrieval tasks spanning a variety of domains. The BEIR benchmark used the Normalized Cumulative Discount Gain (nDCG@10) metric. As shown in Table \ref{tab:res_retrieval}, both of our sentence embedding models significantly outperform the baselines on the \nasair{} task while still maintaining good performance on several of the BEIR tasks (individual results on BEIR tasks shown in Appendix \ref{app:beir-results}). Notably, \cassretsmall{} outperformed \cassretbase{}, on both \nasair{} and BEIR, while being about 4.6x faster.



\section{Industrial Applications of \cassiopea{}}

We show industrial applications of \cassiopea{} models for downstream tasks in the scientific domain.


\subsection{Retrieval and Vector Search}

\textsc{nasa} developed the Science Discovery Engine (\sde{})\footnote{https://sciencediscoveryengine.nasa.gov/}, a search capability that enables the discovery of open data, software and documentation across astrophysics, biological and physical Sciences, Earth science, heliophysics and planetary science \cite{bugbee2022selecting}. To improve search performance, we developed a document retrieval and extractive QA pipeline using the finetuned \cassiopea{} models, with the following components:
\begin{itemize}[topsep=0pt, itemsep=0pt, leftmargin=.2in, parsep=0pt]
    \item \textbf{Sentence Embedding Model:} We use \cassretbase{} to encode a corpus into a vector database, enabling the retrieval of relevant documents based on a user query.
    \item \textbf{Document Re-Ranker Model:} To further improve the relevancy of search results, the retrieved documents are ranked using a document re-ranker model \cassrank{}\footnote{https://huggingface.co/nasa-impact/nasa-smd-ibm-ranker}. This model is fine-tuned from \cassbase{} on the MS-MARCO dataset \cite{bajaj2016ms}.
    \item \textbf{Extractive QA Model:} Answers are extracted using a QA model finetuned from \cassbase{}.
\end{itemize}
This system is expected to be live by mid-December 2024.

\begin{table}[t!]
\begin{adjustbox}{width=0.5\textwidth}
\begin{tabular}{c|c|c}
\hline
 & \robertabase{}& \cassbase{} \\
\hline
\textsc{ms-marco} (MRR@5) & 35.9 & \textbf{36.4}\\
\nasaqa{}  (MRR@5)&  31.1 & \textbf{33.2} \\
\hline
\end{tabular}
\end{adjustbox}
\caption{MRR@5 on re-ranking \nasaqa{} and MS-MARCO tasks using rerankers finetuned from different base models.}
\label{tab:reranker}
\end{table}

\begin{table}[t!]
\begin{adjustbox}{width=0.5\textwidth}
\begin{tabular}{c|c|c|c}
\hline
\textbf{Model} & \multicolumn{2}{c|}{\textbf{Document Retrieval Score}} & \textbf{Answer Quality} \\
& MRR@1 & MRR@3 & Avg. Quality Score \\
\hline
\robertabase{} & 0.54 & 0.62 & 0.60 \\
\cassbase{} & \textbf{0.69} & \textbf{0.78} & \textbf{0.88} \\
\hline
\end{tabular}
\end{adjustbox}
\caption{Avg. Document Retrieval and Answer Quality Scores for 26 questions formulated by experts across astrophysics, biology \& physical science, Earth science, heliophysics \& planetary science domains.}
\label{tab:applications}
\end{table}



 
 First, we compare the performance of \cassrank{} to an identical re-ranker finetuned from \robertabase{} in Table \ref{tab:reranker}. Here, we measure MRR@5 of correctly ranking the most relevant paragraph for the given question. While the \cassrank{} has comparable performance to the \roberta{}-reranker on the MS-Marco dev set, it significantly outperforms the latter on the \nasaqa{} dataset, alluding to better domain contextualization of the \cassbase{} model. 

We then evaluated the end-to-end performance of the domain-adapted model verses the generic \roberta{} model in the aforementioned pipeline. Both systems were queried with a set of questions spanning various thematic areas, and then manually scored by human annotators based on the document relevance and correctness of the extracted answers. 
For assessing document retrieval quality, we use the \textbf{MRR@1} and \textbf{MRR@3} metric, which computes the average reciprocal rank of the highest ranked document from the system's top-1 and top-3 retrieved documents respectively.  For answer quality, experts mark an \textbf{Answer Quality Score}. A score of 1 indicates the correct answer is returned within the first three snippets (a contiguous chunk from the document), 0.5 indicates that the answer is returned in more than three snippets, and 0 indicates no relevant answer is returned. 
Table \ref{tab:applications} shows the superior scores when using \cassiopea{} models, most likely due to the overlap in domain and verbiage of the content indexed by the \sde{} and training corpus of \cassiopea{} models. Example responses from both systems, and a screenshot of the system is shown in Appendix \ref{app:indus-applications}.

\begin{figure}[t!]
    \centering
    \includegraphics[width=1\columnwidth, height=6.8cm]{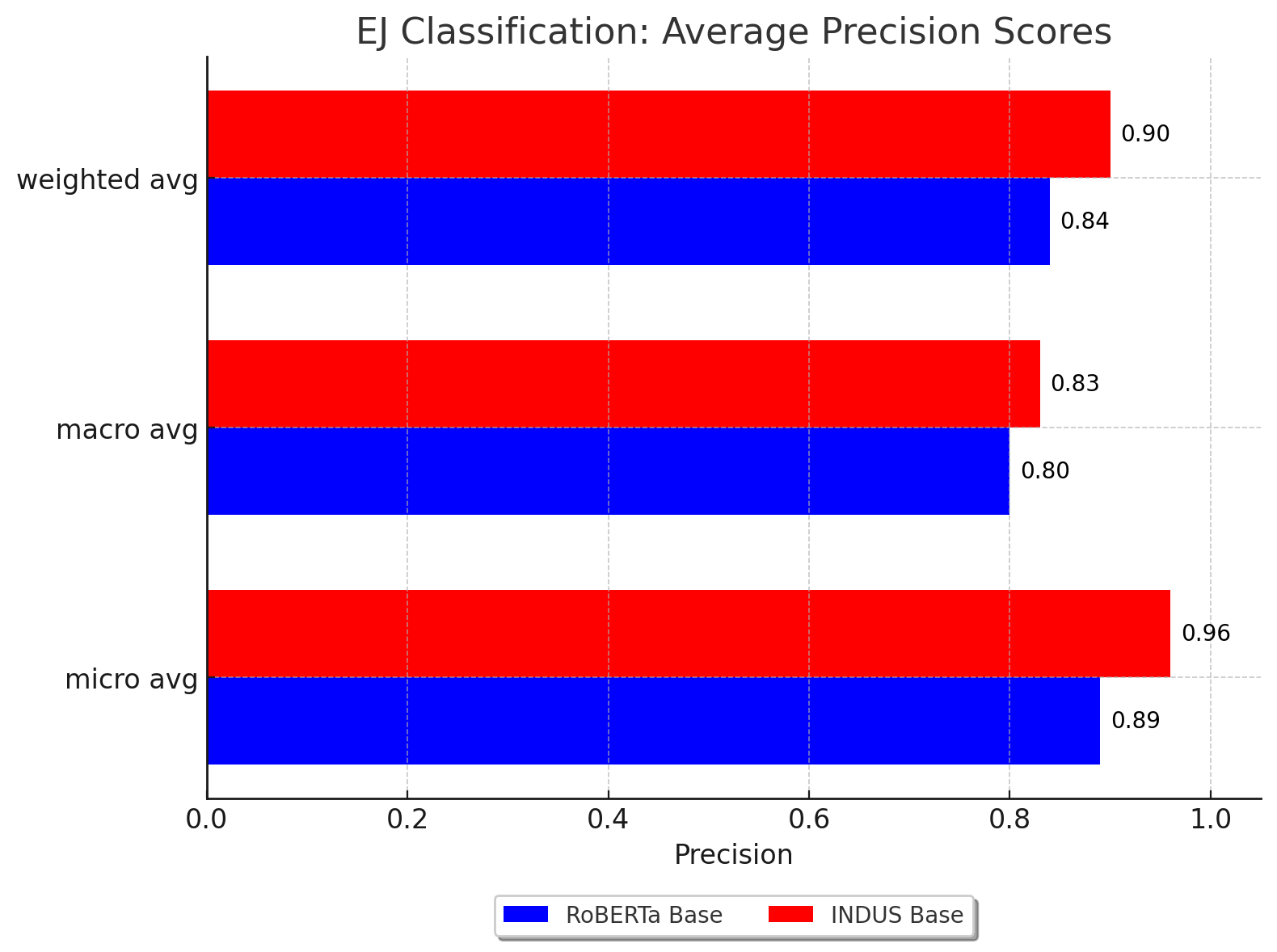}
    \caption{ Average Precision Scores of the \textsc{ej} Indicators Classification Test Set.}
    \label{fig:ej-perf}
\end{figure}

\subsection{Automated Content Curation}
\paragraph{Environmental Justice Portal in \sde{}}

Content curation is a crucial step in providing a high-quality search experience the \sde{}, where Subject Matter Experts (SMEs) identify scientifically relevant information to make available for search and discovery. \cassiopea{} models are being used to automate this time consuming process, for example to identify datasets for specialized search applications like the Environmental Justice Data Search Interface\footnote{https://sciencediscoveryengine.nasa.gov/app/nasa-sba-ej/\#/ej/home}, which focuses on data and metadata related to environmental justice (\textsc{ej}). SMEs identified relevant \textsc{ej} datasets and tagged them with eight indicators: \textit{Human Dimensions, Health \& Air Quality, Climate Change, Food Availability, Disasters, Urban Flooding, Extreme Heat, Water Availability}. This resulted in 139 classification samples which was used to finetune \cassbase{} to develop the multi-label classifier, \ejclass{}\footnote{https://huggingface.co/nasa-impact/ej-classification}. We also added another "Not-EJ" class to identify documents that are not related to \textsc{ej}. This classifier is being used to identify relevant \textsc{ej} content from the \sde{} (live by mid-December 2024). To evaluate model performance, we used a held-out test set comprising 20\% of the 139 samples, stratified equally across all indicators.  As shown in Figure \ref{fig:ej-perf}, the domain-specific model fine-tuned from \cassbase{} has higher precision than the general-purpose model \robertabase{}.


\begin{figure}[t!]
    \centering
    \includegraphics[width=1\columnwidth, height=6.8cm]{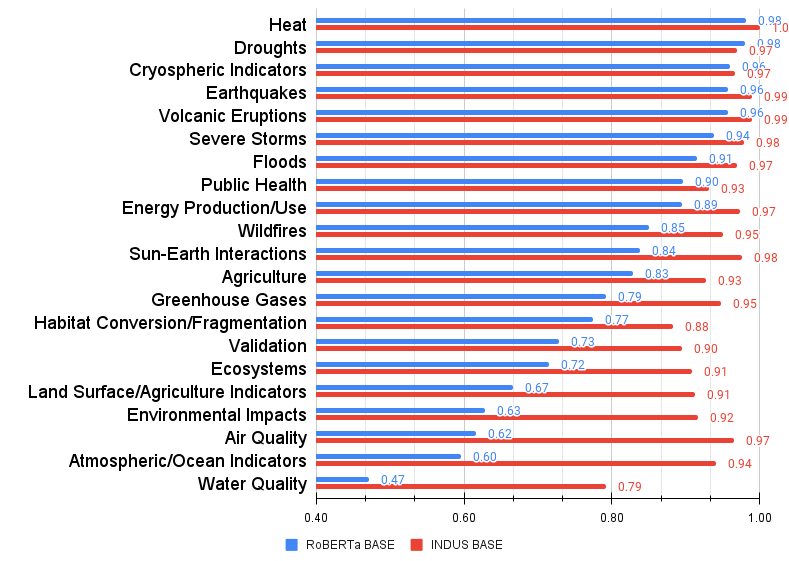}
    \caption{F1-Scores of the classes (GCMD Applied Research Areas) over 1036 test samples.}
    \label{fig:class-perf}
\end{figure}

\paragraph{GCMD Applied Research Area Tags}


Beyond \sde{}, we apply \cassbase{} to categorize scientific publications into 21 Applied Research Areas from the Global Change Master Directory (\citet{gcmd}), as part of a collection that cites datasets from NASA's Goddard Earth Sciences Data and Information Services Center (GES-DISC) and have been annotated by experts. Each publication is annotated with multiple applied research areas allowing for multi-label classification, as detailed in \citet{gerasimov2024bridging}. \cassbase{} was finetined to categorize scientific texts into the aforementioned categories, and is used to enhance publication and dataset discovery in \citet{gesdisc_publications}.  We evaluate the model's performance on 1036 unseen publications, and show in Figure \ref{fig:class-perf} that \cassbase{} outperforms finetuned \robertabase{} by 16\% in terms of macro average F1 score.





\section{Conclusion}
In this work, we presented \cassiopea{}, a constellation of models for use in the science domain and show their applications in industrial settings. We demonstrated the effectiveness of a custom tokenizer and in-domain data for training high-quality encoder models and sentence embedding models. Further, we created smaller versions of the models suitable for applications with latency or resource constraints through state-of-the-art knowledge distillation techniques. For the benefit of the scientific community, we have released all models and benchmarks.



\section*{Acknowledgements}
This work is supported by NASA Grant 80MSFC22M004. We thank all the SMEs who contributed towards the datasets introduced in the paper. We also thank American Geophysical Union (AGU) and the American Meteorological Society (AMS) for providing scientific papers and articles to help build the corpus for the pre-training the \cassbase{} model. We thank the ML team at Sinequa for providing assistance and expertise in finetuning the INDUS models for prototyping in the SDE. We also acknowledge support from the IBM Research AI Hardware Center and the Center for Computational Innovation (CCI) at Rensselaer Polytechnic Institute for computational resources on the AiMOS Supercomputer.

\bibliography{custom}

\begin{thebibliography}{54}
\expandafter\ifx\csname natexlab\endcsname\relax\def\natexlab#1{#1}\fi

\bibitem[{Bajaj et~al.(2016)Bajaj, Campos, Craswell, Deng, Gao, Liu, Majumder, McNamara, Mitra, Nguyen, Rosenberg, Song, Stoica, Tiwary, and Wang}]{bajaj2016ms}
Payal Bajaj, Daniel Campos, Nick Craswell, Li~Deng, Jianfeng Gao, Xiaodong Liu, Rangan Majumder, Andrew McNamara, Bhaskar Mitra, Tri Nguyen, Mir Rosenberg, Xia Song, Alina Stoica, Saurabh Tiwary, and Tong Wang. 2016.
\newblock \href {http://arxiv.org/abs/1611.09268} {Ms marco: A human generated machine reading comprehension dataset}.

\bibitem[{Beltagy et~al.(2019)Beltagy, Lo, and Cohan}]{beltagy-etal-2019-scibert}
Iz~Beltagy, Kyle Lo, and Arman Cohan. 2019.
\newblock \href {https://doi.org/10.18653/v1/D19-1371} {{S}ci{BERT}: A pretrained language model for scientific text}.
\newblock In \emph{Proceedings of the 2019 Conference on Empirical Methods in Natural Language Processing and the 9th International Joint Conference on Natural Language Processing (EMNLP-IJCNLP)}, pages 3615--3620, Hong Kong, China. Association for Computational Linguistics.

\bibitem[{Brown et~al.(2020)Brown, Mann, Ryder, Subbiah, Kaplan, Dhariwal, Neelakantan, Shyam, Sastry, Askell, Agarwal, Herbert-Voss, Krueger, Henighan, Child, Ramesh, Ziegler, Wu, Winter, Hesse, Chen, Sigler, Litwin, Gray, Chess, Clark, Berner, McCandlish, Radford, Sutskever, and Amodei}]{brown-2020-gpt}
Tom Brown, Benjamin Mann, Nick Ryder, Melanie Subbiah, Jared~D Kaplan, Prafulla Dhariwal, Arvind Neelakantan, Pranav Shyam, Girish Sastry, Amanda Askell, Sandhini Agarwal, Ariel Herbert-Voss, Gretchen Krueger, Tom Henighan, Rewon Child, Aditya Ramesh, Daniel Ziegler, Jeffrey Wu, Clemens Winter, Chris Hesse, Mark Chen, Eric Sigler, Mateusz Litwin, Scott Gray, Benjamin Chess, Jack Clark, Christopher Berner, Sam McCandlish, Alec Radford, Ilya Sutskever, and Dario Amodei. 2020.
\newblock \href {https://proceedings.neurips.cc/paper_files/paper/2020/file/1457c0d6bfcb4967418bfb8ac142f64a-Paper.pdf} {Language models are few-shot learners}.
\newblock In \emph{Advances in Neural Information Processing Systems}, volume~33, pages 1877--1901. Curran Associates, Inc.

\bibitem[{Bugbee et~al.(2022)Bugbee, Ramachandran, Acharya, That, Hedman, Eleish, Driessnack, Adams, and Foshee}]{bugbee2022selecting}
Kaylin Bugbee, Rahul Ramachandran, Ashish Acharya, Dai-Hai~Ton That, John Hedman, Ahmed Eleish, Charles Driessnack, Wesley Adams, and Emily Foshee. 2022.
\newblock Selecting approaches for enabling enterprise data search: Nasa's science mission directorate (smd) catalog.
\newblock In \emph{IGARSS 2022-2022 IEEE International Geoscience and Remote Sensing Symposium}, pages 6836--6839. IEEE.

\bibitem[{Clement et~al.(2019)Clement, Bierbaum, O'Keeffe, and Alemi}]{clement2019arxiv}
Colin~B. Clement, Matthew Bierbaum, Kevin~P. O'Keeffe, and Alexander~A. Alemi. 2019.
\newblock \href {http://arxiv.org/abs/1905.00075} {On the use of arxiv as a dataset}.

\bibitem[{Cohan et~al.(2020)Cohan, Feldman, Beltagy, Downey, and Weld}]{specter2020cohan}
Arman Cohan, Sergey Feldman, Iz~Beltagy, Doug Downey, and Daniel~S. Weld. 2020.
\newblock {SPECTER: Document-level Representation Learning using Citation-informed Transformers}.
\newblock In \emph{ACL}.

\bibitem[{Devlin et~al.(2019)Devlin, Chang, Lee, and Toutanova}]{devlin-etal-2019-bert}
Jacob Devlin, Ming-Wei Chang, Kenton Lee, and Kristina Toutanova. 2019.
\newblock \href {https://doi.org/10.18653/v1/N19-1423} {{BERT}: Pre-training of deep bidirectional transformers for language understanding}.
\newblock In \emph{Proceedings of the 2019 Conference of the North {A}merican Chapter of the Association for Computational Linguistics: Human Language Technologies, Volume 1 (Long and Short Papers)}, pages 4171--4186, Minneapolis, Minnesota. Association for Computational Linguistics.

\bibitem[{Doğan et~al.(2014)Doğan, Leaman, and Lu}]{NCBI}
Rezarta~Islamaj Doğan, Robert Leaman, and Zhiyong Lu. 2014.
\newblock \href {https://doi.org/https://doi.org/10.1016/j.jbi.2013.12.006} {Ncbi disease corpus: a resource for disease name recognition and concept normalization}.
\newblock \emph{Journal of Biomedical Informatics}.

\bibitem[{Dunn et~al.(2017)Dunn, Sagun, Higgins, Guney, Cirik, and Cho}]{dunn2017searchqa}
Matthew Dunn, Levent Sagun, Mike Higgins, V.~Ugur Guney, Volkan Cirik, and Kyunghyun Cho. 2017.
\newblock \href {http://arxiv.org/abs/1704.05179} {Searchqa: A new q\&a dataset augmented with context from a search engine}.

\bibitem[{Fader et~al.(2014)Fader, Zettlemoyer, and Etzioni}]{fader_2014_wikianswers}
Anthony Fader, Luke Zettlemoyer, and Oren Etzioni. 2014.
\newblock \href {https://doi.org/10.1145/2623330.2623677} {Open question answering over curated and extracted knowledge bases}.
\newblock In \emph{Proceedings of the 20th ACM SIGKDD International Conference on Knowledge Discovery and Data Mining}, KDD '14, page 1156–1165, New York, NY, USA. Association for Computing Machinery.

\bibitem[{Gao et~al.(2021)Gao, Yao, and Chen}]{gao-etal-2021-simcse}
Tianyu Gao, Xingcheng Yao, and Danqi Chen. 2021.
\newblock \href {https://doi.org/10.18653/v1/2021.emnlp-main.552} {{S}im{CSE}: Simple contrastive learning of sentence embeddings}.
\newblock In \emph{Proceedings of the 2021 Conference on Empirical Methods in Natural Language Processing}, pages 6894--6910, Online and Punta Cana, Dominican Republic. Association for Computational Linguistics.

\bibitem[{{GCMD}()}]{gcmd}
{GCMD}.
\newblock Global change master directory (gcmd).
\newblock \url{https://catalog.data.gov/dataset/global-change-master-directory-gcmd}.
\newblock Accessed: 8 October 2024.

\bibitem[{Gerasimov et~al.(2024)Gerasimov, Savtchenko, Alfred, Acker, Wei, and Binita}]{gerasimov2024bridging}
Irina Gerasimov, Andrey Savtchenko, Jerome Alfred, James Acker, Jennifer Wei, and KC~Binita. 2024.
\newblock Bridging the gap: Enhancing prominence and provenance of nasa datasets in research publications.
\newblock \emph{Data Science Journal}, 23(1).

\bibitem[{{GES-DISC Portal}()}]{gesdisc_publications}
{GES-DISC Portal}.
\newblock Nasa publications.
\newblock \url{https://disc.gsfc.nasa.gov/information/publications}.
\newblock Accessed: 8 October 2024.

\bibitem[{Gu et~al.(2021)Gu, Tinn, Cheng, Lucas, Usuyama, Liu, Naumann, Gao, and Poon}]{gu2022blurb}
Yu~Gu, Robert Tinn, Hao Cheng, Michael Lucas, Naoto Usuyama, Xiaodong Liu, Tristan Naumann, Jianfeng Gao, and Hoifung Poon. 2021.
\newblock \href {https://doi.org/10.1145/3458754} {Domain-specific language model pretraining for biomedical natural language processing}.
\newblock \emph{ACM Trans. Comput. Healthcare}, 3(1).

\bibitem[{Hasibi et~al.(2017)Hasibi, Nikolaev, Xiong, Balog, Bratsberg, Kotov, and Callan}]{dbpedia}
Faegheh Hasibi, Fedor Nikolaev, Chenyan Xiong, Krisztian Balog, Svein~Erik Bratsberg, Alexander Kotov, and Jamie Callan. 2017.
\newblock \href {https://doi.org/10.1145/3077136.3080751} {Dbpedia-entity v2: A test collection for entity search}.
\newblock In \emph{Proceedings of the 40th International ACM SIGIR Conference on Research and Development in Information Retrieval}, SIGIR '17, page 1265–1268, New York, NY, USA. Association for Computing Machinery.

\bibitem[{Hinton et~al.(2014)Hinton, Vinyals, and Dean}]{hinton2014distilling}
Geoffrey Hinton, Oriol Vinyals, and Jeff Dean. 2014.
\newblock {Distilling the Knowledge in a Neural Network}.
\newblock In \emph{NeurIPS Deep Learning Worksop}.

\bibitem[{Hong et~al.(2023)Hong, Ajith, Pauloski, Duede, Chard, and Foster}]{hong2023diminishing}
Zhi Hong, Aswathy Ajith, Gregory Pauloski, Eamon Duede, Kyle Chard, and Ian Foster. 2023.
\newblock \href {http://arxiv.org/abs/2205.11342} {The diminishing returns of masked language models to science}.

\bibitem[{Huang and Cole(2022)}]{huang2022batterybert}
Shu Huang and Jacqueline~M Cole. 2022.
\newblock \href {https://doi.org/10.1021/acs.jcim.2c00035} {Batterybert: A pretrained language model for battery database enhancement}.
\newblock \emph{J. Chem. Inf. Model.}, page DOI: 10.1021/acs.jcim.2c00035.

\bibitem[{Karpukhin et~al.(2020)Karpukhin, Oguz, Min, Lewis, Wu, Edunov, Chen, and Yih}]{karpukhin-etal-2020-dense}
Vladimir Karpukhin, Barlas Oguz, Sewon Min, Patrick Lewis, Ledell Wu, Sergey Edunov, Danqi Chen, and Wen-tau Yih. 2020.
\newblock \href {https://doi.org/10.18653/v1/2020.emnlp-main.550} {Dense passage retrieval for open-domain question answering}.
\newblock In \emph{Proceedings of the 2020 Conference on Empirical Methods in Natural Language Processing (EMNLP)}, pages 6769--6781, Online. Association for Computational Linguistics.

\bibitem[{Khosla et~al.(2020)Khosla, Teterwak, Wang, Sarna, Tian, Isola, Maschinot, Liu, and Krishnan}]{khosla_contrastive_2020}
Prannay Khosla, Piotr Teterwak, Chen Wang, Aaron Sarna, Yonglong Tian, Phillip Isola, Aaron Maschinot, Ce~Liu, and Dilip Krishnan. 2020.
\newblock \href {https://proceedings.neurips.cc/paper_files/paper/2020/file/d89a66c7c80a29b1bdbab0f2a1a94af8-Paper.pdf} {Supervised contrastive learning}.
\newblock In \emph{Advances in Neural Information Processing Systems}, volume~33, pages 18661--18673. Curran Associates, Inc.

\bibitem[{Kinney et~al.(2023)Kinney, Anastasiades, Authur, Beltagy, Bragg, Buraczynski, Cachola, Candra, Chandrasekhar, Cohan, Crawford, Downey, Dunkelberger, Etzioni, Evans, Feldman, Gorney, Graham, Hu, Huff, King, Kohlmeier, Kuehl, Langan, Lin, Liu, Lo, Lochner, MacMillan, Murray, Newell, Rao, Rohatgi, Sayre, Shen, Singh, Soldaini, Subramanian, Tanaka, Wade, Wagner, Wang, Wilhelm, Wu, Yang, Zamarron, Van~Zuylen, and Weld}]{kinney-etal-2023-s2odp}
Rodney Kinney, Chloe Anastasiades, Russell Authur, Iz~Beltagy, Jonathan Bragg, Alexandra Buraczynski, Isabel Cachola, Stefan Candra, Yoganand Chandrasekhar, Arman Cohan, Miles Crawford, Doug Downey, Jason Dunkelberger, Oren Etzioni, Rob Evans, Sergey Feldman, Joseph Gorney, David Graham, Fangzhou Hu, Regan Huff, Daniel King, Sebastian Kohlmeier, Bailey Kuehl, Michael Langan, Daniel Lin, Haokun Liu, Kyle Lo, Jaron Lochner, Kelsey MacMillan, Tyler Murray, Chris Newell, Smita Rao, Shaurya Rohatgi, Paul Sayre, Zejiang Shen, Amanpreet Singh, Luca Soldaini, Shivashankar Subramanian, Amber Tanaka, Alex~D. Wade, Linda Wagner, Lucy~Lu Wang, Chris Wilhelm, Caroline Wu, Jiangjiang Yang, Angele Zamarron, Madeleine Van~Zuylen, and Daniel~S. Weld. 2023.
\newblock \href {https://doi.org/10.48550/ARXIV.2301.10140} {The semantic scholar open data platform}.

\bibitem[{Kwiatkowski et~al.(2019)Kwiatkowski, Palomaki, Redfield, Collins, Parikh, Alberti, Epstein, Polosukhin, Devlin, Lee, Toutanova, Jones, Kelcey, Chang, Dai, Uszkoreit, Le, and Petrov}]{kwiatkowski-etal-2019-natural}
Tom Kwiatkowski, Jennimaria Palomaki, Olivia Redfield, Michael Collins, Ankur Parikh, Chris Alberti, Danielle Epstein, Illia Polosukhin, Jacob Devlin, Kenton Lee, Kristina Toutanova, Llion Jones, Matthew Kelcey, Ming-Wei Chang, Andrew~M. Dai, Jakob Uszkoreit, Quoc Le, and Slav Petrov. 2019.
\newblock {Natural Questions: A Benchmark for Question Answering Research}.
\newblock \emph{Transactions of the ACL}.

\bibitem[{Lee et~al.(2019)Lee, Yoon, Kim, Kim, Kim, So, and Kang}]{lee-2019-biobert}
Jinhyuk Lee, Wonjin Yoon, Sungdong Kim, Donghyeon Kim, Sunkyu Kim, Chan~Ho So, and Jaewoo Kang. 2019.
\newblock \href {https://doi.org/10.1093/bioinformatics/btz682} {{BioBERT: a pre-trained biomedical language representation model for biomedical text mining}}.
\newblock \emph{Bioinformatics}, 36(4):1234--1240.

\bibitem[{Lewis et~al.(2020)Lewis, Liu, Goyal, Ghazvininejad, Mohamed, Levy, Stoyanov, and Zettlemoyer}]{lewis-etal-2020-bart}
Mike Lewis, Yinhan Liu, Naman Goyal, Marjan Ghazvininejad, Abdelrahman Mohamed, Omer Levy, Veselin Stoyanov, and Luke Zettlemoyer. 2020.
\newblock \href {https://doi.org/10.18653/v1/2020.acl-main.703} {{BART}: Denoising sequence-to-sequence pre-training for natural language generation, translation, and comprehension}.
\newblock In \emph{Proceedings of the 58th Annual Meeting of the Association for Computational Linguistics}, pages 7871--7880, Online. Association for Computational Linguistics.

\bibitem[{Lewis et~al.(2021)Lewis, Wu, Liu, Minervini, Küttler, Piktus, Stenetorp, and Riedel}]{lewis_paq_2021}
Patrick Lewis, Yuxiang Wu, Linqing Liu, Pasquale Minervini, Heinrich Küttler, Aleksandra Piktus, Pontus Stenetorp, and Sebastian Riedel. 2021.
\newblock \href {https://doi.org/10.1162/tacl_a_00415} {{PAQ: 65 Million Probably-Asked Questions and What You Can Do With Them}}.
\newblock \emph{Transactions of the Association for Computational Linguistics}, 9:1098--1115.

\bibitem[{Li et~al.(2016)Li, Sun, Johnson, Sciaky, Wei, Leaman, Davis, Mattingly, Wiegers, and Lu}]{Li2016BioCreativeVC}
Jiao Li, Yueping Sun, Robin~J. Johnson, Daniela Sciaky, Chih-Hsuan Wei, Robert Leaman, Allan~Peter Davis, Carolyn~J. Mattingly, Thomas~C. Wiegers, and Zhiyong Lu. 2016.
\newblock \href {https://api.semanticscholar.org/CorpusID:88817} {Biocreative v cdr task corpus: a resource for chemical disease relation extraction}.
\newblock \emph{Database: The Journal of Biological Databases and Curation}, 2016.

\bibitem[{Li et~al.(2023)Li, Zhang, Zhang, Long, Xie, and Zhang}]{li2023general}
Zehan Li, Xin Zhang, Yanzhao Zhang, Dingkun Long, Pengjun Xie, and Meishan Zhang. 2023.
\newblock \href {http://arxiv.org/abs/2308.03281} {Towards general text embeddings with multi-stage contrastive learning}.

\bibitem[{Liu et~al.(2019)Liu, Ott, Goyal, Du, Joshi, Chen, Levy, Lewis, Zettlemoyer, and Stoyanov}]{liu2019roberta}
Yinhan Liu, Myle Ott, Naman Goyal, Jingfei Du, Mandar Joshi, Danqi Chen, Omer Levy, Mike Lewis, Luke Zettlemoyer, and Veselin Stoyanov. 2019.
\newblock \href {https://arxiv.org/abs/1907.11692} {Roberta: A robustly optimized bert pretraining approach}.
\newblock \emph{arXiv preprint arXiv:1907.11692}.

\bibitem[{Lo et~al.(2020)Lo, Wang, Neumann, Kinney, and Weld}]{lo-etal-2020-s2orc}
Kyle Lo, Lucy~Lu Wang, Mark Neumann, Rodney Kinney, and Daniel Weld. 2020.
\newblock \href {https://doi.org/10.18653/v1/2020.acl-main.447} {{S}2{ORC}: The semantic scholar open research corpus}.
\newblock In \emph{Proceedings of the 58th Annual Meeting of the Association for Computational Linguistics}, pages 4969--4983, Online. Association for Computational Linguistics.

\bibitem[{Maia et~al.(2018)Maia, Handschuh, Freitas, Davis, McDermott, Zarrouk, and Balahur}]{fiqa}
Macedo Maia, Siegfried Handschuh, Andr\'{e} Freitas, Brian Davis, Ross McDermott, Manel Zarrouk, and Alexandra Balahur. 2018.
\newblock \href {https://doi.org/10.1145/3184558.3192301} {Www'18 open challenge: Financial opinion mining and question answering}.
\newblock In \emph{Companion Proceedings of the The Web Conference 2018}, WWW '18, page 1941–1942, Republic and Canton of Geneva, CHE. International World Wide Web Conferences Steering Committee.

\bibitem[{Nentidis et~al.(2020)Nentidis, Bougiatiotis, Krithara, and Paliouras}]{Nentidis_2020}
Anastasios Nentidis, Konstantinos Bougiatiotis, Anastasia Krithara, and Georgios Paliouras. 2020.
\newblock \href {https://doi.org/10.1007/978-3-030-43887-6_51} {\emph{Results of the Seventh Edition of the BioASQ Challenge}}, page 553–568. Springer International Publishing.

\bibitem[{Radford et~al.(2019)Radford, Wu, Child, Luan, Amodei, and Sutskever}]{radford2019gpt2}
Alec Radford, Jeff Wu, Rewon Child, David Luan, Dario Amodei, and Ilya Sutskever. 2019.
\newblock \href {https://d4mucfpksywv.cloudfront.net/better-language-models/language_models_are_unsupervised_multitask_learners.pdf} {Language models are unsupervised multitask learners}.

\bibitem[{Raffel et~al.(2020)Raffel, Shazeer, Roberts, Lee, Narang, Matena, Zhou, Li, and Liu}]{raffel-2020-t5}
Colin Raffel, Noam Shazeer, Adam Roberts, Katherine Lee, Sharan Narang, Michael Matena, Yanqi Zhou, Wei Li, and Peter~J. Liu. 2020.
\newblock Exploring the limits of transfer learning with a unified text-to-text transformer.
\newblock \emph{Journal of Machine Learning Research}, 21(1).

\bibitem[{Rajpurkar et~al.(2018)Rajpurkar, Jia, and Liang}]{sq2}
Pranav Rajpurkar, Robin Jia, and Percy Liang. 2018.
\newblock \href {http://arxiv.org/abs/1806.03822} {Know what you don't know: Unanswerable questions for squad}.
\newblock \emph{CoRR}, abs/1806.03822.

\bibitem[{Rajpurkar et~al.(2016)Rajpurkar, Zhang, Lopyrev, and Liang}]{rajpurkar2016squad}
Pranav Rajpurkar, Jian Zhang, Konstantin Lopyrev, and Percy Liang. 2016.
\newblock {SQuAD: 100,000+ Questions for Machine Comprehension of Text}.
\newblock In \emph{EMNLP}.

\bibitem[{Reimers and Gurevych(2019)}]{reimers-gurevych-2019-sentence}
Nils Reimers and Iryna Gurevych. 2019.
\newblock \href {https://doi.org/10.18653/v1/D19-1410} {Sentence-{BERT}: Sentence embeddings using {S}iamese {BERT}-networks}.
\newblock In \emph{Proceedings of the 2019 Conference on Empirical Methods in Natural Language Processing and the 9th International Joint Conference on Natural Language Processing (EMNLP-IJCNLP)}, pages 3982--3992, Hong Kong, China. Association for Computational Linguistics.

\bibitem[{Thakur et~al.(2021)Thakur, Reimers, Rücklé, Srivastava, and Gurevych}]{n2021beir}
Nandan Thakur, Nils Reimers, Andreas Rücklé, Abhishek Srivastava, and Iryna Gurevych. 2021.
\newblock \href {http://arxiv.org/abs/2104.08663} {Beir: A heterogenous benchmark for zero-shot evaluation of information retrieval models}.

\bibitem[{Thorne et~al.(2018)Thorne, Vlachos, Christodoulopoulos, and Mittal}]{thorne-etal-2018-fever}
James Thorne, Andreas Vlachos, Christos Christodoulopoulos, and Arpit Mittal. 2018.
\newblock \href {https://doi.org/10.18653/v1/N18-1074} {{FEVER}: a large-scale dataset for fact extraction and {VER}ification}.
\newblock In \emph{Proceedings of the 2018 Conference of the North {A}merican Chapter of the Association for Computational Linguistics: Human Language Technologies, Volume 1 (Long Papers)}, pages 809--819, New Orleans, Louisiana. Association for Computational Linguistics.

\bibitem[{Tinn et~al.(2023)Tinn, Cheng, Gu, Usuyama, Liu, Naumann, Gao, and Poon}]{tinn2023fine}
Robert Tinn, Hao Cheng, Yu~Gu, Naoto Usuyama, Xiaodong Liu, Tristan Naumann, Jianfeng Gao, and Hoifung Poon. 2023.
\newblock Fine-tuning large neural language models for biomedical natural language processing.
\newblock \emph{Patterns}, 4(4).

\bibitem[{Touvron et~al.(2023)Touvron, Lavril, Izacard, Martinet, Lachaux, Lacroix, Rozière, Goyal, Hambro, Azhar, Rodriguez, Joulin, Grave, and Lample}]{touvron2023llama}
Hugo Touvron, Thibaut Lavril, Gautier Izacard, Xavier Martinet, Marie-Anne Lachaux, Timothée Lacroix, Baptiste Rozière, Naman Goyal, Eric Hambro, Faisal Azhar, Aurelien Rodriguez, Armand Joulin, Edouard Grave, and Guillaume Lample. 2023.
\newblock \href {http://arxiv.org/abs/2302.13971} {Llama: Open and efficient foundation language models}.

\bibitem[{Trivedi et~al.(2023)Trivedi, Udagawa, Merler, Panda, El-Kurdi, and Bhattacharjee}]{trivedi2023neural}
Aashka Trivedi, Takuma Udagawa, Michele Merler, Rameswar Panda, Yousef El-Kurdi, and Bishwaranjan Bhattacharjee. 2023.
\newblock Neural architecture search for effective teacher-student knowledge transfer in language models.
\newblock \emph{arXiv preprint arXiv:2303.09639}.

\bibitem[{Tsatsaronis et~al.(2015)Tsatsaronis, Balikas, Malakasiotis, Partalas, Zschunke, Alvers, Weissenborn, Krithara, Petridis, Polychronopoulos, Almirantis, Pavlopoulos, Baskiotis, Gallinari, Artiéres, Ngomo, Heino, Gaussier, Barrio-Alvers, Schroeder, and Androutsopoulos}]{bioasq_retrieval}
George Tsatsaronis, Georgios Balikas, Prodromos Malakasiotis, Ioannis Partalas, Matthias Zschunke, Michael~R. Alvers, Dirk Weissenborn, Anastasia Krithara, Sergios Petridis, Dimitris Polychronopoulos, Yannis Almirantis, John Pavlopoulos, Nicolas Baskiotis, Patrick Gallinari, Thierry Artiéres, Axel-Cyrille~Ngonga Ngomo, Norman Heino, Eric Gaussier, Liliana Barrio-Alvers, Michael Schroeder, and Ion Androutsopoulos. 2015.
\newblock \href {https://doi.org/https://doi.org/10.1186/s12859-015-0564-6} {An overview of the bioasq large-scale biomedical semantic indexing and question answering competition}.
\newblock \emph{BMC Bioinformatics}.

\bibitem[{Udagawa et~al.(2023)Udagawa, Trivedi, Merler, and Bhattacharjee}]{udagawa-etal-2023-comparative}
Takuma Udagawa, Aashka Trivedi, Michele Merler, and Bishwaranjan Bhattacharjee. 2023.
\newblock \href {https://doi.org/10.18653/v1/2023.emnlp-industry.3} {A comparative analysis of task-agnostic distillation methods for compressing transformer language models}.
\newblock In \emph{Proceedings of the 2023 Conference on Empirical Methods in Natural Language Processing: Industry Track}, pages 20--31, Singapore. Association for Computational Linguistics.

\bibitem[{van~den Oord et~al.(2019)van~den Oord, Li, and Vinyals}]{oord2019representation}
Aaron van~den Oord, Yazhe Li, and Oriol Vinyals. 2019.
\newblock \href {http://arxiv.org/abs/1807.03748} {Representation learning with contrastive predictive coding}.

\bibitem[{Vaswani et~al.(2017)Vaswani, Shazeer, Parmar, Uszkoreit, Jones, Gomez, Kaiser, and Polosukhin}]{vaswani_attention}
Ashish Vaswani, Noam Shazeer, Niki Parmar, Jakob Uszkoreit, Llion Jones, Aidan~N Gomez, \L~ukasz Kaiser, and Illia Polosukhin. 2017.
\newblock \href {https://proceedings.neurips.cc/paper_files/paper/2017/file/3f5ee243547dee91fbd053c1c4a845aa-Paper.pdf} {Attention is all you need}.
\newblock In \emph{Advances in Neural Information Processing Systems}, volume~30. Curran Associates, Inc.

\bibitem[{Wadden et~al.(2020)Wadden, Lin, Lo, Wang, van Zuylen, Cohan, and Hajishirzi}]{wadden-etal-2020-fact}
David Wadden, Shanchuan Lin, Kyle Lo, Lucy~Lu Wang, Madeleine van Zuylen, Arman Cohan, and Hannaneh Hajishirzi. 2020.
\newblock \href {https://doi.org/10.18653/v1/2020.emnlp-main.609} {Fact or fiction: Verifying scientific claims}.
\newblock In \emph{Proceedings of the 2020 Conference on Empirical Methods in Natural Language Processing (EMNLP)}, pages 7534--7550, Online. Association for Computational Linguistics.

\bibitem[{Walker et~al.(2021)Walker, Trewartha, Huo, Lee, Cruse, Dagdelen, Dunn, Persson, Ceder, and Jain}]{walker2021impact}
Nicholas Walker, Amalie Trewartha, Haoyan Huo, Sanghoon Lee, Kevin Cruse, John Dagdelen, Alexander Dunn, Kristin Persson, Gerbrand Ceder, and Anubhav Jain. 2021.
\newblock The impact of domain-specific pre-training on named entity recognition tasks in materials science.
\newblock \emph{Available at SSRN 3950755}.

\bibitem[{Wang et~al.(2022)Wang, Yang, Huang, Jiao, Yang, Jiang, Majumder, and Wei}]{wang2022text}
Liang Wang, Nan Yang, Xiaolong Huang, Binxing Jiao, Linjun Yang, Daxin Jiang, Rangan Majumder, and Furu Wei. 2022.
\newblock \href {http://arxiv.org/abs/2212.03533} {Text embeddings by weakly-supervised contrastive pre-training}.

\bibitem[{Wang et~al.(2021)Wang, Bao, Huang, Dong, and Wei}]{wang-etal-2021-minilmv2}
Wenhui Wang, Hangbo Bao, Shaohan Huang, Li~Dong, and Furu Wei. 2021.
\newblock \href {https://doi.org/10.18653/v1/2021.findings-acl.188} {{M}ini{LM}v2: Multi-head self-attention relation distillation for compressing pretrained transformers}.
\newblock In \emph{Findings of the Association for Computational Linguistics: ACL-IJCNLP 2021}, pages 2140--2151, Online. Association for Computational Linguistics.

\bibitem[{Xiao et~al.(2022)Xiao, Liu, Shao, and Cao}]{xiao-etal-2022-retromae}
Shitao Xiao, Zheng Liu, Yingxia Shao, and Zhao Cao. 2022.
\newblock \href {https://doi.org/10.18653/v1/2022.emnlp-main.35} {{R}etro{MAE}: Pre-training retrieval-oriented language models via masked auto-encoder}.
\newblock In \emph{Proceedings of the 2022 Conference on Empirical Methods in Natural Language Processing}, pages 538--548, Abu Dhabi, United Arab Emirates. Association for Computational Linguistics.

\bibitem[{Xiao et~al.(2023)Xiao, Liu, Zhang, and Muennighoff}]{xiao2023cpack}
Shitao Xiao, Zheng Liu, Peitian Zhang, and Niklas Muennighoff. 2023.
\newblock \href {http://arxiv.org/abs/2309.07597} {C-pack: Packaged resources to advance general chinese embedding}.

\bibitem[{Xu et~al.(2023)Xu, Shao, Chen, and Liu}]{xu-etal-2023-distillcse}
Jiahao Xu, Wei Shao, Lihui Chen, and Lemao Liu. 2023.
\newblock \href {https://doi.org/10.18653/v1/2023.findings-emnlp.547} {{D}istill{CSE}: Distilled contrastive learning for sentence embeddings}.
\newblock In \emph{Findings of the Association for Computational Linguistics: EMNLP 2023}, pages 8153--8165, Singapore. Association for Computational Linguistics.

\bibitem[{Yang et~al.(2018)Yang, Qi, Zhang, Bengio, Cohen, Salakhutdinov, and Manning}]{yang-etal-2018-hotpotqa}
Zhilin Yang, Peng Qi, Saizheng Zhang, Yoshua Bengio, William Cohen, Ruslan Salakhutdinov, and Christopher~D. Manning. 2018.
\newblock \href {https://doi.org/10.18653/v1/D18-1259} {{H}otpot{QA}: A dataset for diverse, explainable multi-hop question answering}.
\newblock In \emph{Proceedings of the 2018 Conference on Empirical Methods in Natural Language Processing}, pages 2369--2380, Brussels, Belgium. Association for Computational Linguistics.

\end{thebibliography}

\appendix

\section{Training Details: Encoder Models}
\label{app:train-encoder}
\cassbase{} was trained with the masked language modeling objective, using the default hyperparameters recommended in Table 9 of \citet{liu2019roberta}.We change the effective batch size to $9216$, training for 500K steps on 192 V100 GPUs.

\casssmall{} was distilled using the MiniLMv2 approach \cite{wang-etal-2021-minilmv2}, with an effective batch size of 480 for 500K steps 
on 30 V100 GPUs.

\section{Sentence Embedding Training Data}
\label{app:embedding-data}

Table~\ref{tab:embedding-data} shows the various data sources used for training embedding models. All data is presented in the form of text-pairs, where each item in the pair may be a sentence or a paragraph.  We used about 360 million pairs for training and used in-batch negatives.

\begin{table*}[t]
\centering
\begin{adjustbox}{width=1\textwidth}
\begin{tabular}{llll}
\hline
Dataset & Num. Pairs & Data Category & Data Format \\
\hline
StackOverflow$^\dagger$ & 18562443 & Title-Body & s2p \\
StackExchange Math$^\dagger$ & 2201906 & Title-Body & s2p \\
S2ORC {[}title - abstract{]} \cite{lo-etal-2020-s2orc}& 41769185 & Title-Body & s2p \\
S2ORC Citation Pairs {[}Abstracts{]} \cite{lo-etal-2020-s2orc} & 52603982 & Title-Body & p2p \\
StackExchange {[}title - body{]}$^\dagger$ & 5415570 & Title-Body & s2p \\
Wikipedia \cite{fader_2014_wikianswers} & 6458670 & Title-Body & s2p \\
Arxiv \cite{clement2019arxiv} & 2358545 & Title-Body & s2p \\
NASA ADS {[}title - abstract{]} (\S\ref{sec:data}) & 2633240 & Title-Body & s2p \\
PubMed {[}title - abstract{]} (\S\ref{sec:data}) & 24001387 & Title-Body & s2p \\
PMC {[}title - abstract{]} (\S\ref{sec:data}) & 2585537 & Title-Body & s2p \\
\hline
StackExchange Duplicate Questions {[}title-body - title-body{]}$^\dagger$ & 250460 & Duplicate Questions & p2p \\
StackExchange Duplicate Questions {[}body - body{]}$^\dagger$ & 250519 & Duplicate Questions & p2p \\
StackExchange Duplicate Questions {[}title - title{]}$^\dagger$ & 304525 & Duplicate Questions & s2s \\
WikiAnswer Pairs \cite{fader_2014_wikianswers} & 77427422 & Duplicate Questions & s2s \\
\hline
Specter Pairs \cite{specter2020cohan} & 684100 & Citation Pairs & s2s \\
S2ORC Citation Pairs {[}Titles{]} \cite{lo-etal-2020-s2orc} & 52603982 & Citation Pairs & s2s \\
\hline
SQuAD \cite{rajpurkar2016squad}& 87599 & Question Answers & s2p \\
NQ \cite{kwiatkowski-etal-2019-natural}& 100231 & Question Answers & s2p \\
SearchQA \cite{dunn2017searchqa}& 582261 & Question Answers & s2p \\
StackExchange {[}title - answer{]}$^\dagger$ & 4067139 & Question Answers & s2p \\
StackExchange {[}title-body - answer{]}$^\dagger$ & 187195 & Question Answers & p2p \\
PAQ  \cite{lewis_paq_2021} & 64371441 & Question Answers & s2p \\
\hline
FEVER \cite{thorne-etal-2018-fever}$^\ast$ & 109810 & Fact Verification & s2p \\
HotpotQA \cite{yang-etal-2018-hotpotqa}$^\ast$ & 85000 & Question Answering & s2p \\
\hline

\end{tabular}
\end{adjustbox}
\caption{Training Data for Embedding Models. The training data totals to around 360M pairs. Data Format denotes s2p for sentence-to-paragraph mappings, s2s for sentence-to-sentence mappings, and p2p for paragraph-to-paragraph mappings. $^\dagger$Downloaded from https://huggingface.co/datasets/flax-sentence-embeddings/stackexchange\_xml. $^\ast$Only used for Distillation.}  
\label{tab:embedding-data}
\end{table*}

\section{Training Details: Sentence Embedding}
\label{app:sent-embed-train}

For the base retriever model, we use the following loss: for a triple $\{q,p^+, P^-\}$ of a query, a relevant (positive) passage, and a set of non-relevant (negative) passages $P^- = {\{p^-_j\}_{j=1}^{m}}$, We define the InfoNCE loss \cite{oord2019representation} as:

\begin{equation}
    \mathcal{L}_{IC} = -\frac{1}{n} \sum_{i=1}^{n} \mathrm{log} \frac{e^{s(q_i, p_i^+)}}{Z_i}
\label{eq:cont-loss}
\end{equation}
\begin{equation}
\begin{aligned}
    Z_i = \sum_{j} e^{s(q_i, p_j)} + \sum_{j} e^{s(q_j, p_i^+)} \\ 
    + \sum_{j \neq i } e^{s(q_i, q_j)} + \sum_{j \neq i } e^{s(p_i^+, p_j)^-}
\end{aligned}
\end{equation}


where $s(q, p)$ is a measure of temperature-scaled cosine similarity between the embeddings of query and a passage measured by (where $\mathbf{E}(\cdot)$ denotes the embedding function and $\tau$ is the temperature):
\begin{equation}
    s(q,p) = \frac{1}{\tau} \frac{\mathbf{E}(q) \cdot\mathbf{E}(p)} {\|\mathbf{E}(q)\|\|\mathbf{E}(p)\|}
\label{eq:sim-score}
\end{equation}

We trained each stage on 2 A100 GPUs with an effective batch size of 1,024. We first trained with unsupervised data for 300K steps followed by an additional 100K steps with the supervised data. We used a learning rate of $2e-5$ and $\tau=0.02$ during both these steps.

We used knowledge distillation techniques introduced in \cite{xu-etal-2023-distillcse} to create a smaller, more efficient retriever (\cassretsmall{}) through the supervision of \cassretbase{}.  Specifically, for a sentence $x_i$ and its corresponding in-batch element pairs $\{x_i, x_j\}_{j=1,j \neq i}^{m}$, we minimized the cross entropy between the teacher's distribution $p_t$ of similarity scores between pairs and the student's distribution, $p_s$. Following \citet{hinton2014distilling}, we also scaled the output distribution of both teacher and student by a temperature, $\tau_{KD}$:
\begin{equation}
    \mathcal{L}_{KD} = - \sum_{i=1}^{n} \sum_{j=1}^{m} p_t(x_i, x_j) \mathrm{log} p_s(x_i, x_j)
\end{equation}
\begin{equation}
    p_s(x_i, x_j) =  \frac{e^{s_s(x_i, x_j)/\tau_{KD}}}{\sum_{k=1}^{m} e^{s_s(x_i, x_k)/\tau_{KD}}} 
\end{equation}
\begin{equation}
    p_t(x_i, x_j) = \frac{e^{s_t(x_i, x_j)/\tau_{KD}}}{\sum_{k=1}^{m} e^{s_t(x_i, x_k)/\tau_{KD}}}
\end{equation}

Here, $s_s(x_i, x_j)$ and $s_t(x_i, x_j)$ represent the similarity scores between two pairs $\{x_i, x_j\}$, defined in Equation \ref{eq:sim-score} for the student and teacher respectively. Note, $\tau_{KD}$ is the distillation temperature and is unrelated to the distance-temperature $\tau$ defined in Equation \ref{eq:sim-score}.

For the Retro-MAE style pretraining \cite{xiao-etal-2022-retromae}, we trained on 8 A100 GPUs with an effective batch size of 128 for 2 epochs with a learning rate of $2e-5$. For the stage-wise distillation, we trained on 2 A100 GPUs for 300K steps with an effective batch size of 2,048, and learning rate of $7e-4$. Through experimentation, we found that $\tau_{KD} = 4$ performed the best, and we keep $\tau=0.02$ as in the non-distilled case.

\section{Size of Proposed Benchmarks}
\label{app:benchmark-details}

The aim of our benchmark is to measure performance of models on three important yet orthogonal natural language understanding tasks, namely Named Entity Recognition, Extractive Question Answering and Information Retrieval. Each task further focuses on a different subset of domains of interest, specifically including those which are not covered by existing tests. 

Moreover, we believe the size of each dataset to be comparable to other widely used domain-specific test sets in IR (eg. num. queries in BioASQ \cite{bioasq_retrieval}, FiQA \cite{fiqa}, DBPedia \cite{dbpedia} and SciFact \cite{wadden-etal-2020-fact} tasks from BEIR), QA (eg. num. questions in BioASQ \cite{Nentidis_2020} from BLURB ), and NER (eg. num. entities in NCBI-disease \cite{NCBI}, BC5-Chem \cite{Li2016BioCreativeVC}, BC5-Disease \cite{Li2016BioCreativeVC} from BLURB). We hope that the introduction of these datasets will serve as a much needed first step towards advancing benchmarking capabilities in this important field.

\section{Ablation Study: Stage-wise Distillation for Embedding Model}

\label{app:embedding-stagewise-ablation}

For the distilled embedding models, we find that stage-wise distillation does not benefit performance as much as a one-step process, combining all the supervised and unsupervised data. As shown in Table \ref{tab:ablation}, the stage-wise approach underperformed the one-stage approach by 1 percentage point for both \nasair{} and on BEIR.

\begin{table}[h]
\centering
\begin{adjustbox}{width=0.5\textwidth}
\begin{tabular}{l|l|c|c}
\hline
Model & Training & \nasair{} & BEIR Avg. \\
 \hline
\cassretsmall{} & One-Stage & 0.73 & 0.42 \\
\cassretsmall{} & Stagewise & 0.72 & 0.41 \\
\hline
\end{tabular}
\end{adjustbox}
\caption{Ablation Study: Evaluation results on \nasair{} and BEIR. \nasair{} showed Recall\@10 while BEIR reported nDCG\@10.}
\label{tab:ablation}
\end{table}

\section{Complete Results on BEIR Benchmark}
\label{app:beir-results}
Table \ref{tab:res_retrieval_full} shows the per-dataset results on the BEIR tasks.

\begin{table*}[t]
\centering
\begin{adjustbox}{width=1.0\textwidth}
\begin{tabular}{l|lllllllllllll}
\hline
Model &  \multicolumn{13}{c}{BEIR Eval}  \\
\hline
 &   TREC- & NFCorpus & NQ & HotPotQA & FiQA & ArguaAna & Touche & DBPedia & Scidocs & FEVER & Climate  & SciFact &  AVG. \\
  &   Covid &  &  &  &  &  &  &  &  &  & FEVER &  & BEIR  \\
 \hline
\robertabase{} & 0.47 & 0.30 & 0.54 & 0.34 & 0.38 & 0.52 & 0.18 & 0.25 & 0.22 & 0.46 & 0.14 & 0.67 & 0.37 \\
\textsc{bge}$_\textsc{base}$  & 0.78 & 0.37 & 0.54 & 0.73 & 0.41 & 0.64 & 0.26 & 0.41 & 0.22 & 0.86 & 0.31 & 0.74 & 0.52 \\
\cassretbase{}  & 0.56 & 0.32 & 0.54 & 0.49 & 0.36 & 0.54 & 0.17 & 0.31 & 0.21 & 0.56 & 0.14 & 0.74 & 0.41 \\
\hline
\textsc{minilm-v2}  & 0.47 & 0.32 & 0.44 & 0.47 & 0.35 & 0.50 & 0.17 & 0.32 & 0.22 & 0.52 & 0.25 & 0.65 & 0.39 \\
\textsc{bge}$_\textsc{small}$ & 0.76	& 0.34 &0.50 & 0.70 & 0.40 & 0.60 &	0.26 & 0.40& 0.21& 0.87& 0.32 & 0.71	& 0.51 \\
\cassretsmall{}  & 0.55 & 0.31 & 0.53 & 0.48 & 0.29 & 0.50 & 0.21 & 0.33 & 0.23 & 0.61 & 0.23 & 0.71 & 0.42 \\
\hline
\end{tabular}
\end{adjustbox}
\caption{Evaluation results BEIR.}
\label{tab:res_retrieval_full}
\end{table*}

\section{Applications of \cassiopea{} for Retrieval: Performance and Interface}
\label{app:indus-applications}

\begin{figure*}[t!]
\centering
\includegraphics[width=\textwidth]{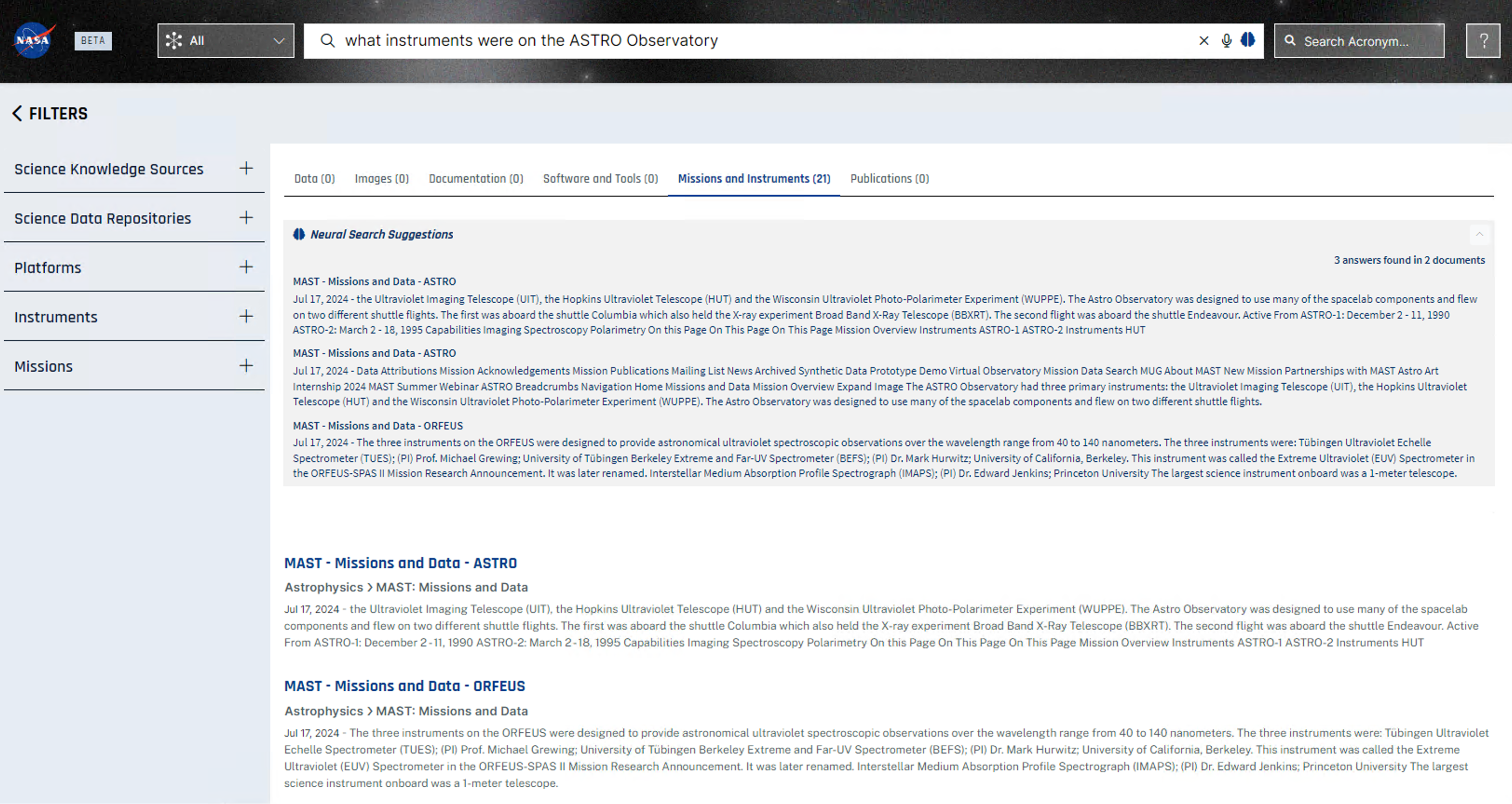}
\caption{Interface to the Information Retrieval System built with \cassiopea{}. A user searches for a query and obtains snippets extracted from the document that contain relevant information, along with a list of relevant documents from which these snippets are extracted (screenshot edited to protect anonymity).}
\label{fig-indus-application-search}
\end{figure*}  

We show the interface for the Science Discovery Engine, the information retrieval system built with \cassbase{} in Figure \ref{fig-indus-application-search}, showing retrieved documents relevant to the seach query along with snippets with pertinant information. 

Table \ref{tab:roberta} and Table \ref{tab:cass} contain a few sample queries created for benchmarking by a human evaluator to compare the performance of the knowledge retrieval system leveraging \cassbase{} finetuned models with the one using generic \robertabase{} model. As shown, \cassbase{} usually provides a higher document and answer quality.

\begin{table*}[t]
\centering
\begin{adjustbox}{width=1.00\textwidth}
\begin{tabular}{|p{4cm}|p{4cm}|p{2cm}|p{10cm}|p{1.2cm}|}
\hline
Question &  Document Title & Retrieved Document Rank & Retrieved Document & Answer Quality Score \\
\hline
What does MODIS measure? & The MODIS Near-IR Water Vapor Algorithm & 3 & MODIS is a major facility instrument on the EOS polar orbiting satellite platforms (Asrar and Greenstone, 1995; King et al., 1992; Salomonson et al., 1989) designed to measure biological and physical  processes on a global scale every 1 to 2 days. It is a 36-channel scanning  radiometer covering the spectral region 0.4 - 15 $\mu$m. Five near-IR MODIS channels are useful for remote sensing of water vapor. & 0.5 \\
\hline
Which algorithm document describes the ZAVG product? & CERES ATBD Subsystem 8.0 - Monthly Regional, Zonal, and Global &
1 & Compute Regional, Zonal and Global Averages (Subsystem 8.0) This appendix describes the data products which are produced by the algorithms in this subsystem. The table below summarizes these products, listing the CERES and EOSDIS product  codes or abbreviations, a short product name, the product type, the production frequency, and volume estimates for each individual product as well as a complete data month of production. The product types are defined as follows:  Archival products: &
  0.5 \\ \hline
Where did Perseverance land on Mars? & None & No relevant document retrieved & Perseverance's First Autonav Drive  This image was taken during the first drive of NASA's Perseverance rover on Mars on March 4, 2021. Perseverance landed on Feb. 18, 2021, and the team has been spending the weeks since landing check... Perseverance Is Roving on Mars  This map shows where NASA's Perseverance Mars rover will be dropping 10 samples that a future mission could pick up. A Map of Perseverance's Depot Samples  This image is an edited version of the last  360-degree panorama taken by the Opportunity rover's Pancam from May 13 through June 10, 2018. & 0.0 \\ \hline
At what point in space is the JWST located? & \#JwstArt Juried Art Show & 1 & Lines depict the direction of the waves reaching the telescope's instruments. Heat waves depicted highlight the temperature difference between the two sides of the solar shield. In order to analyze infrared light, the JWST needs to operate at 50 Kelvin (-223C/-370F) because the heat from the sun can interfere with the data entering the  instruments. The bottom portion shows the relative location of the telescope after launch just outside earth umbra at the L2 Point about 1.5 million km from Earth. & 1.0 \\ \hline
What is the data policy for JWST? & Quick Start Guide - MAST Docs - STScI Outerspace & No Relevant Document Retrieved & No Answer Found & 0.0 \\ \hline
\end{tabular}
\end{adjustbox}
\caption{Sample Questions from Human Evaluation of Vector Search pipeline leveraging \robertabase{} model.}
\label{tab:roberta}
\end{table*}

\begin{table*}[t]
\centering
\begin{adjustbox}{width=1.0\textwidth}
\begin{tabular}{|p{4cm}|p{4cm}|p{2cm}|p{10cm}|p{1.2cm}|}
\hline
Question &  Document Title & Retrieved Document Rank & Retrieved Document & Answer Quality Score \\
\hline
What does MODIS measure? & DRAFT OF THE MODIS LEVEL 1B ATBD version 2.0 (ATBMOD - 01) & 1 & The MODIS raw output is a small, rapidly varying signal superimposed on a large background that varies more slowly, due to the thermal  drifts and $1/f$ noise.  Like its predecessor instruments,  MODIS views space as its background subtraction reference and a full-aperture blackbody as its  second reference for calibration. MODIS measures space and blackbody reference before and after  each Earth view scan line. If $1/f$ noise is known at the time MODIS is viewing the space and blackbody reference then $1/f$ noise in the Earth view sector can be interpolated between four known & 1.0 \\ \hline
Which algorithm document describes the ZAVG product?& CERES ATBD Subsystem 8.0 Monthly Regional, Zonal, and Global & 1  & 
Monthly Zonal and Global Radiative Fluxes and Clouds (ZAVG). The Monthly Zonal and Global Radiative Fluxes and Clouds (ZAVG) product is a summary of  the zonal and global averages of the radiative fluxes and cloud properties, probably most suitable for  inclusion in the Earth. Observing System Data  and Information System (EOSDIS) Information Management System (IMS) as a browse product. This product is the CERES equivalent to the zonal  averages and global averages in the ERBE S-4 product.  ZAVG is an archival product  produced by the TISA subsystem for each instrument and for each  combination of instruments. &  0.5 \\ \hline
Where did Perseverance land on Mars? & Sample Tube 266 - NASA Mars Exploration & 1 & Perseverance will land at the Red Planet's Jezero Crater a little after 3:40 p.m. EST (12:40 p.m. PST) on Feb...  Perseverance on Mars  NASA's Perseverance Mars rover is using its self-driving capabilities as it treks across Jezero Crater seeking signs of ancient life and gathering rock and soil samples for planned return to Earth.  How Perseverance Drives on Mars  This high-resolution image shows one of the six wheels aboard NASA's Perseverance Mars rover, which landed on Feb.18, 2021. The image was taken by one of Perseverance's color Hazard Cameras & 1.0 \\ \hline
At what point in space is the JWST located? & \#JwstArt Juried Art Show Webb/NASA & 3 & None & 1.0 \\ \hline
What is the data policy for JWST? & Solar System Observation FAQ For Scientists Webb/NASA & 1 & The JWST Science \& Operations Center will be located at the Space Telescope Science Institute (STScI) in Baltimore, MD. Competition will be fierce!  What is my proprietary time?  The baseline period for exclusive access to your JWST data is one year, as for HST and other missions. Some types of programs will have a shorter or zero exclusive access period. Proposers can also voluntarily reduce or waive their proprietary data rights. After the end of the exclusive access period the observations will be available for archival research. &
  1.0 \\ 
\hline
\end{tabular}
\end{adjustbox}
\caption{Sample Questions from Human Evaluation of Vector Search pipeline leveraging \cassbase{} model}
\label{tab:cass}
\end{table*}

\end{document}